\documentclass[10pt,twocolumn,letterpaper]{article}

\usepackage{bbding}
\usepackage{algorithm, algpseudocode}
\usepackage{bm}
\usepackage{multirow}
\usepackage[pagenumbers]{cvpr} % To force page numbers, e.g. for an arXiv version
\usepackage{multirow}
\usepackage{diagbox}
\usepackage{fontawesome}
\usepackage[accsupp]{axessibility}
\usepackage[table ]{xcolor}
\newcommand{\vect}[1]{\boldsymbol{\mathbf{#1}}}
\definecolor{cvprblue}{rgb}{0.21,0.49,0.74}
\usepackage{xcolor}         % colors
\definecolor{LightCyan}{RGB}{232,241,255}
\usepackage[pagebackref,breaklinks,colorlinks,citecolor=cvprblue]{hyperref}

\title{Towards Transferable Targeted 3D Adversarial Attack in the Physical World}

\makeatletter
\def\thanks#1{\protected@xdef\@thanks{\@thanks
        \protect\footnotetext{#1}}}
\makeatother

\author{Yao Huang$^{1,3}$, Yinpeng Dong$^{2, 3\dagger}$, Shouwei Ruan$^{1}$, Xiao Yang$^{2}$, Hang Su$^{2}$, Xingxing Wei$^{1\dagger}$\thanks{$^\dagger$Corresponding authors.} \\
$^{1}$ Institute of Artificial Intelligence, Beihang University, Beijing 100191, China\\
$^{2}$ Dept. of Comp. Sci. and Tech., Institute for AI, Tsinghua-Bosch Joint ML Center,\\
THBI Lab, BNRist Center,
Tsinghua University, Beijing, 100084, China\hspace{2ex}$^{3}$ RealAI\\
\tt\small{\{y\_huang, shouweiruan, xxwei\}@buaa.edu.cn}\hspace{2ex}
\tt\small{\{dongyinpeng, suhangss\}@mail.tsinghua.edu.cn}\\
\tt\small{\{yangxiao19\}@mails.tsinghua.edu.cn} 
}

\begin{document}
\maketitle

\begin{abstract}
Compared with transferable untargeted attacks, transferable targeted adversarial attacks could specify the misclassification categories of adversarial samples, posing a greater threat to security-critical tasks. In the meanwhile, 3D adversarial samples, due to their potential of multi-view robustness, can more comprehensively identify weaknesses in existing deep learning systems, possessing great application value. However, the field of transferable targeted 3D adversarial attacks remains vacant. The goal of this work is to develop a more effective technique that could generate transferable targeted 3D adversarial examples, filling the gap in this field. To achieve this goal, we design a novel framework named \textbf{TT3D} that could rapidly reconstruct from few multi-view images into \textbf{T}ransferable \textbf{T}argeted \textbf{3D} textured meshes. While existing mesh-based texture optimization methods compute gradients in the high-dimensional mesh space and easily fall into local optima, leading to unsatisfactory transferability and distinct distortions, TT3D innovatively performs dual optimization towards both feature grid and Multi-layer Perceptron (MLP) parameters in the grid-based NeRF space, which significantly enhances black-box transferability while enjoying naturalness. Experimental results show that TT3D not only exhibits superior cross-model transferability but also maintains considerable adaptability across different renders and vision tasks. More importantly, we produce 3D adversarial examples with 3D printing techniques in the real world and verify their robust performance under various scenarios. Code is publicly available at: \url{hhttps://github.com/Aries-iai/TT3D}
\vspace{-3ex}
\end{abstract}    
\section{Introduction}
\label{sec:intro}
Despite the unprecedented performance of deep neural networks (DNNs) in numerous tasks, including image classification~\cite{rawat2017deep} and object detection~\cite{zhao2019object, bo2021ship}, these models are vulnerable to adversarial examples~\cite{szegedy2013intriguing, goodfellow2014explaining, moosavi2016deepfool, madry2017towards, dong2018boosting, Xie_2019_CVPR, thys2019fooling, 9779913, 9999043, dong2022viewfool, Wei_2023_CVPR, ruan2023improving, Wei_2023_ICCV, wei2023unified, wei2023infrared}. These adversarial examples are shown to have good transferability across models \cite{Liu2016, dong2018boosting, Xie_2019_CVPR}, emerging as a key concern since they can evade black-box models in practical applications. The majority of transferable attacks are untargeted, aiming to induce misclassification of the target model. However, transferable targeted attacks, which mislead DNNs to produce predetermined misclassifications, can lead to a more severe threat in real-world applications.
Crafting such transferable targeted attacks is particularly challenging because they not only need to achieve a specific misclassification but also must avoid overfitting to ensure effective attacks across various models, which requires further investigation. 

\begin{figure}[!t]
  \centering
  \includegraphics[width=\linewidth]{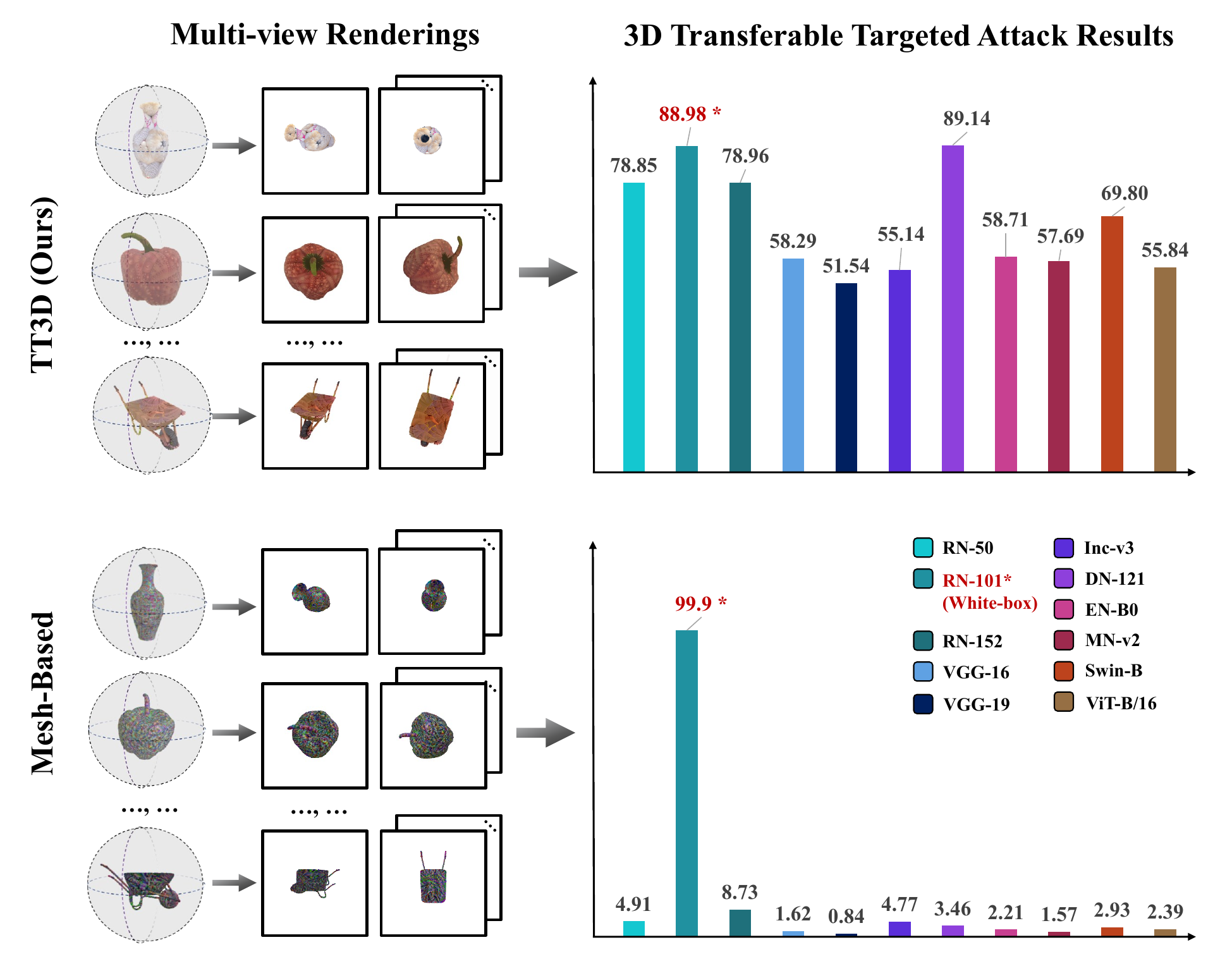}
   \caption{A comparison of transferable targeted attack performance in the 3D domain between our TT3D and the enhanced version of the typical mesh-based optimization method~\cite{xiao2019meshadv}, as detailed in \cref{basic}. The surrogate model is ResNet-101~\cite{He_2016_CVPR} and we can see TT3D shows remarkable transferability.}
   \label{fig:example}
\vspace{-3.5ex}
\end{figure}

\begin{table*}[t]
\small 
% \footnotesize
\setlength{\tabcolsep}{2.5pt}
\begin{center}
\resizebox{\linewidth}{!}{
\begin{tabular}{c|cccccccc}
\toprule
& AdvTurtle~\cite{athalye2018synthesizing} & Zeng\etal~\cite{zeng2019adversarial} & MeshAdv~\cite{xiao2019meshadv} & Oslund\etal~\cite{oslund2022multiview} & FCA~\cite{wang2022fca}  & DTA~\cite{suryanto2022dta} & ACTIVE~\cite{suryanto2023active} & \textbf{TT3D (Ours)} \\
\midrule
Target Task & Classify  & Classify & Classify & Classify & Detect & Detect & Detect+Segment$^{*}$  & Classify+Detect$^{*}$+Caption$^{*}$\\\hline
Data &  3D model & 3D model & 3D model & 3D model & 3D model &  3D model& 3D model &Few Multi-view images\\\hline
3D Attack Type & Texture & Texture & Geometry/Texture & Texture & Texture & Texture & Texture & Texture+Geometry \\\hline
Targeted &  \faCheck &  \faTimes & \faCheck  & \faTimes & \faTimes & \faTimes &\faTimes & \textbf{\faCheck}\\\hline
Transferability & \faTimes &  \faTimes &  \faTimes & \faCheck &  \faCheck & \faCheck & \faCheck & \faCheck\\\hline
Naturalness & \faTimes &  \faTimes & \faTimes & \faTimes & \faCheck & \faTimes & \faCheck & \textbf{\faCheck}\\\hline
Physical Attack & \faCheck & \faTimes & \faTimes & \faCheck & \faTimes & \faCheck & \faCheck  &\textbf{\faCheck}\\
\bottomrule
\end{tabular}}
\end{center}
\vspace{-3ex}
\caption{A comparison among different 3D attack methods regarding target vision tasks, used data, 3D attack types, transferability, whether conducting targeted attacks, naturalness, and whether conducting attacks in the physical world. $^{*}$ represents the transferred task .}
\vspace{-1.5ex}
\label{tab:related-work}
\end{table*}

While transferable targeted adversarial attacks in the 2D domain have been extensively studied~\cite{li2020towards, naseer2021generating, zhao2021success, wang2023towards}, the exploration of transferable targeted attacks in the 3D domain remains vacant.  
Compared with 2D adversarial attacks, 3D attacks possess greater practical value in the real world due to the potential of consistent attacks across various viewpoints. As illustrated in \cref{tab:related-work}, existing 3D attack methods~\cite{athalye2018synthesizing, zeng2019adversarial, xiao2019meshadv, dong2022isometric, wang2022fca, suryanto2022dta, suryanto2023active} could ensure multi-view effectiveness by modifying pre-existing meshes' textures or geometry due to meshes' 3D consistency. But, they struggle to simultaneously ensure transferability and targeted attack, especially for physical attacks. In addition, it is also difficult to maintain the naturalness of these adversarial examples.

Based on the above discussions, this paper aims to generate \textbf{transferable and natural 3D adversarial examples for physically targeted attacks}.
However, there  exist two challenges to meet our goal: (1) Prior mesh-based optimization methods involve direct alterations to vertex colors in the high-dimensional mesh space, easily falling into the overfitting and thus leading to unsatisfactory transferability. Thus, the first challenge is how to design an optimization method that deviates from the mesh space, avoiding overfitting for better transferability. (2) Existing methods struggle to simultaneously ensure attack performance and naturalness, whose attack process easily accompanies the extremely unnatural phenomena, such as visual anomalies in appearance or distortion or fragmentation of the 3D mesh, \etc. Therefore, how to balance the attack performance and the visual naturalness is another challenge.

To meet the above challenges, we design a novel framework called \textbf{TT3D} that could rapidly reconstruct a transferable targeted 3D adversarial textured mesh with guaranteed naturalness. To be specific, driven by the recent advancement in multi-view reconstruction techniques (\ie,~grid-based NeRF~\cite{muller2022instant}), we first reconstruct an initial 3D mesh from its multi-view images, but instead of directly manipulating in the mesh space, we innovatively perform adversarial fine-tuning of appearance rendering parameters in the grid-based NeRF space. Such a method not only successfully avoids overfitting but also eliminates previous methods' reliance on the pre-existing 3D mesh, greatly lowering the cost. The technical details can be found as follows:

Firstly, to enhance the transferability, we design a dual optimization strategy for adversarial fine-tuning. This optimization strategy simultaneously targets the parameters of the appearance feature grid and the corresponding MLP (feature grids and MLP are both components of grid-based NeRF, detailed in ~\cref{nerf}), thereby effectively perturbing at both the foundational feature level and the more advanced decision-making layers. Additionally, it's worth mentioning that, though solely altering the geometric structure is challenging for achieving transferable targeted attacks, we observe that incorporating changes in geometric information under a texture optimization-focused method, can enhance the performance (shown in \cref{basic}). Thus, we add the optimization of vertex coordinates into our methodology. 

Secondly, to ensure the naturalness of the 3D adversarial textured mesh, we implement constraints on appearance and geometry during optimization. These include measures for visual consistency and vertex proximity, along with specialized loss functions to prevent mesh deformities and ensure surface smoothness. For 3D adversarial objects' robustness in the real world, we also introduce an EOT that could effectively integrate various transformations in both 3D and 2D spaces to more realistically simulate real-world conditions. A comparison of our TT3D's performance in transferable targeted attacks with the typical mesh-based optimization method is shown in ~\cref{fig:example}.

\begin{figure*}[!ht]
    \centering
    \includegraphics[width=0.975\linewidth]{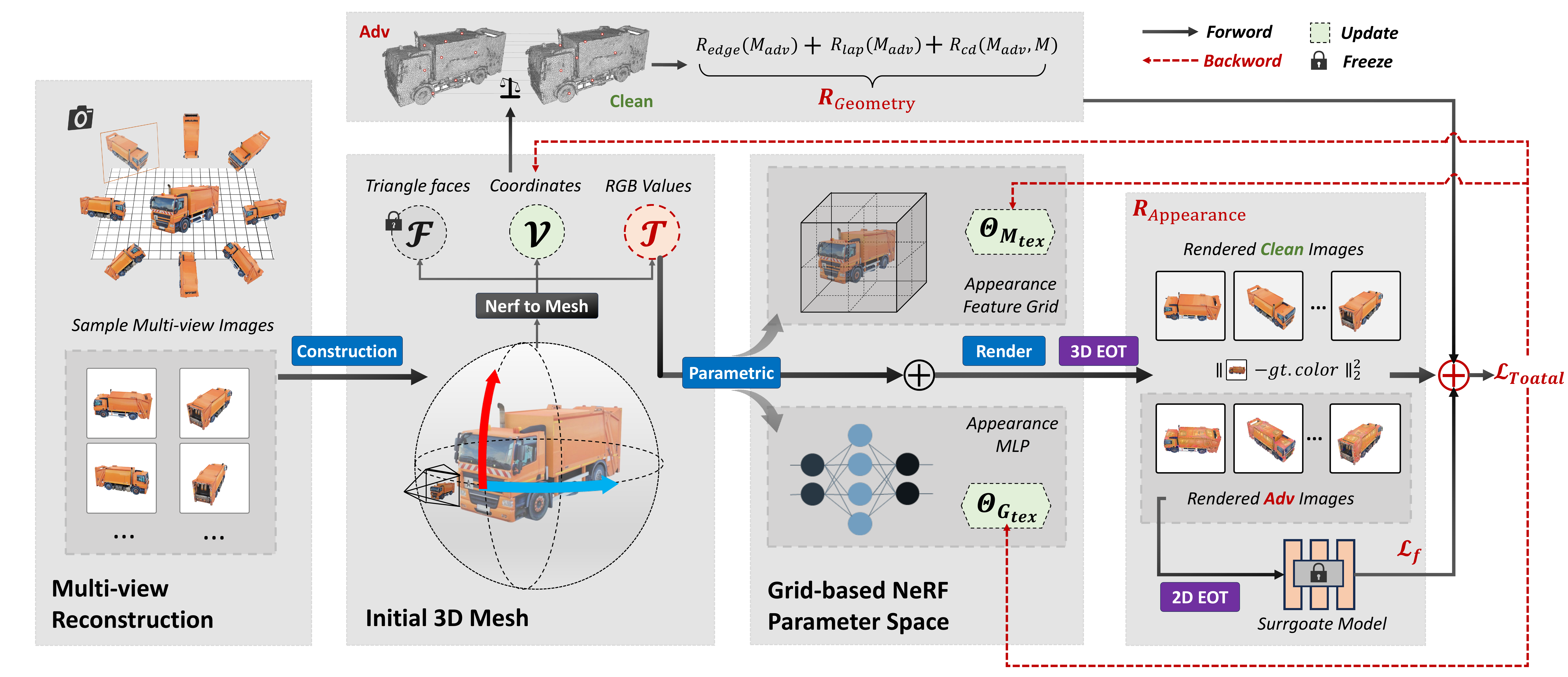}
    \caption{An overview of our TT3D framework. We first utilize 3D multi-view reconstruction technology, \ie, grid-based NeRF with marching cubes techniques to obtain the initial clean 3D mesh. Then, we perform adversarial fine-tuning in the textual parameter space of grid-based NeRF instead of directly altering the texture $\mathcal{T}$, supplemented with geometric perturbations at vertex positions $\mathcal{V}$. To ensure the naturalness simultaneously, we add constraints to the distance between the 3D adversarial samples and the initial ones in terms of both texture and geometric structure when performing optimization .}
    \label{fig:framework3d}
\vspace{-3.5ex}
\end{figure*}
%贡献如下%

The contributions of this paper are as follows:
% \vspace{-0.1ex}
% \begin{itemize}[leftmargin=0.7cm, topsep=0.03cm]
% \setlength{\itemsep}{4.5pt}
\begin{itemize}
    \item We propose a novel framework called \textbf{TT3D} for generating transferable targeted 3D adversarial examples, which is the first work to fill a critical gap in this field and eliminate the previous reliance on pre-existing 3D meshes, expanding the feasible region of 3D attacks.

    \item We design a dual optimization strategy within the grid-based NeRF space, which effectively perturbs at both the foundational feature level and the more sophisticated decision-making echelons of the neural network, and ensures the naturalness in the meanwhile.
    
    \item Experimental results reveal that our 3D adversarial object can easily be misclassified as a given label across various victim models, renders, and tasks. Moreover, we produce 3D adversarial objects in the real world by utilizing 3D printing and validate their robust performance under various settings, like different viewpoints and backgrounds.
\end{itemize}

\section{Related Work}
\label{sec:related}

\subsection{Transferable Targeted Adversarial Attack}
Transferable targeted adversarial attacks in the 2D domain have been extensively studied. For instance, Li \etal~\cite{li2020towards} improve the transferability of targeted attacks by addressing `noise curing' through self-adaptive gradient magnitudes and metric learning. Zhao \etal~\cite{zhao2021success} demonstrate that simple logit loss-based attacks, without extensive resources, can achieve unexpected effectiveness in targeted transferability, challenging traditional, resource-intensive methods. The works of Wang \etal~\cite{wang2023towards} and Naseer \etal~\cite{naseer2021generating} further advance the field, introducing sophisticated approaches for capturing class distributions and aligning image distributions. Moreover, Yang \etal~\cite{yang2022boosting} also propose a hierarchical generative network-based approach for crafting targeted transfer-based adversarial examples, showcasing the potential of generative models in this domain. However, the above works only focus on the 2D domain while the realm of transferable targeted 3D adversarial attacks remains vacant, posing a necessity for investigations.

\subsection{3D Adversarial Attack}
Since Athalye \etal~\cite{athalye2018synthesizing} verified the existence of 3D adversarial examples, plenty of generation methods~\cite{xiao2019meshadv, zeng2019adversarial, oslund2022multiview, wang2022fca, suryanto2022dta, suryanto2023active} have emerged. 
Among them, except for MeshAdv~\cite{xiao2019meshadv}, which has attempted to change the mesh geometry for 3D attacks but still fails to perform transferable targeted attacks, the rest methods~\cite{zeng2019adversarial, oslund2022multiview, wang2022fca, suryanto2022dta, suryanto2023active}, including MeshAdv (optional), are based on modifying the texture of the 3D mesh. We can split texture-based attack methods into two categories. For methods like ~\cite{xiao2019meshadv, zeng2019adversarial, oslund2022multiview}, they directly alter vertex colors in the high-dimensional mesh space, leading to unavoidable overfitting. On the other hand, works like \cite{wang2022fca,suryanto2022dta, suryanto2023active} optimize a texture map and render it onto a specified 3D model (\eg, a car), enhancing transferability to some extent. However, their transferability is limited to non-targeted attacks in object detection, lacking comparability. Moreover, most of these attacks are confined to the digital realm, and physical attacks often come with visual unnaturalness. To address these issues, our work, without the reliance on 3D models, can effectively perform an 3D attack that not only ensures attack performance (\ie, transferable targeted 3D attacks feasible even in real-world scenarios), but also could meet demands for naturalness.
\section{Methodology}
\label{sec:method}
In this section, we detail the proposed TT3D framework for generating transferable targeted 3D adversarial examples, as presented in \cref{fig:framework3d}. TT3D adopts grid-based NeRF \cite{muller2022instant} to model 3D objects from multi-view images and perform optimization. As below, we first introduce the background knowledge of grid-based NeRF and then present the problem formulation and our optimization strategy.
\subsection{Preliminary}
\label{nerf}
Given a set of multi-view images, vanilla NeRF~\cite{martin2021nerf} encodes a real-world object into a continuous volumetric radiance field $F: (\vect{x},\vect{d})\rightarrow (\vect{c}, \sigma)$. $F$ is approximated by a multi-layer perceptron (MLP), which takes a 3D location $\vect{x}\in\mathbb{R}^3$ and a unit-norm viewing direction $\vect{d}\in\mathbb{R}^3$ as inputs, and outputs a volume density $\sigma\in\mathbb{R}^+$ and emitted RGB color $\vect{c}\in[0,1]^3$. Despite its efficacy, vanilla NeRF often faces challenges in terms of computational efficiency.

In addressing the limitations of traditional methods, grid-based NeRF techniques~\cite{muller2022instant, sun2022direct, fridovich2022plenoxels, tang2023delicate} employ two distinct 3D grids, namely $G_{geo}$ and $G_{tex}$, to explicitly represent geometric and textural information of objects. This structured approach, utilizing feature grids, maps a 3D point $\vect{x}$ to corresponding feature vectors $\vect{f_{geo}}$ and $\vect{f_{tex}}$ for both geometry and texture, which are expressed as:
\begin{equation}
\label{map}
\vect{f_{geo}}(\vect{x}) = G_{geo}(\vect{x}; \Theta_{G_{geo}}), \vect{f_{tex}}(\vect{x}) = G_{tex}(\vect{x}; \Theta_{G_{tex}}),
\end{equation}
where $\Theta_{G_{geo}}$ and $\Theta_{G_{tex}}$ denote the sets of feature vectors within $G_{geo}$ and $G_{tex}$, respectively.

To derive the final rendering attributes, \ie, volume density $\sigma$ and emitted color $\vect{c}$, these feature vectors $\vect{f_{geo}}$ and $\vect{f_{tex}}$ are then processed through two shallow MLPs: $M_{geo}$ and $M_{tex}$, formulated as:
\begin{equation}
\sigma \leftarrow M_{geo}(\vect{f_{geo}}(\vect{x}); \Theta_{M_{geo}}), \vect{c} \leftarrow M_{tex}(\vect{f_{tex}}(\vect{x}); \Theta_{M_{tex}}),
\end{equation}
where $\Theta_{M_{geo}}$ and $\Theta_{M_{tex}}$ are weights of $M_{geo}$ and $M_{tex}$.

In this paper, by leveraging grid-based NeRF~\cite{muller2022instant} with Marching Cubes~\cite{lorensen1998marching}, we first efficiently reconstruct 3D mesh as an accurate geometry and continue adversarial finetuning within the parameter space of grid-based NeRF.

\subsection{Problem Formulation}
\label{sec:problem}

For the 3D adversarial attack task, we aim to develop an effective method that can generate transferable targeted 3D adversarial samples and maintain their visual naturalness. Similar to~\cite{athalye2018synthesizing}, we also propose to craft 3D adversarial examples as textured mesh, which can fully leverage 3D printing techniques for physically realizable adversarial attacks.

Specifically, we define the reconstructed mesh representation of the 3D object as $\mathcal{M} =(\mathcal{V},\mathcal{T},\mathcal{F})$,  where $\mathcal{V}\in\mathbb{R}^{n\times 3}$ is the $xyz$ coordinates of $n$ vertices, $\mathcal{T}\in\mathbb{R}^{n\times 3}$ is the $rgb$ value of vertices, and $\mathcal{F}\in\mathbb{Z}^{m\times 3}$ is the set of $m$ triangle faces which encodes each triangle with the indices of vertices.  Here, we do not pose changes to the mesh topology $\mathcal{F}$ to prevent serious distortions of geometric structures, and instead, we turn to $\mathcal{V}$ and $\mathcal{T}$.

In this paper, we focus on the challenging transferable targeted 3D attacks against image classification models under different views. Given a surrogate classifier $f: \mathcal{X}\rightarrow \mathcal{Y}$, the goal of the attack is to generate a 3D adversarial example $\mathcal{M}_{adv}=(\mathcal{V}^{*}, \mathcal{T}^{*}, \mathcal{F})$ for the original one $\mathcal{M}$ with color and vertex perturbations. Then, with a differentiable render function~\cite{Laine2020diffrast} and a random viewpoint $\vect{v}$, we render the  $\mathcal{M}_{adv}$ into the corresponding 2D image $\hat{I}_{\vect{v}}(\mathcal{M}_{adv})$, which can be misclassified by other common classifiers as the target class $y^{*}(\neq y)$. In general, the perturbation should be small to make our 3D adversarial example natural in the physical world. Thus, the optimization problem of crafting 3D adversarial examples can be formulated as
\begin{gather}
\mathop{\text{min}}\limits_{\mathcal{V}^{*}, \mathcal{T}^{*}}\mathbb{E}_{\vect{v}\in V}L_{f}(\hat{I}_{\vect{v}}(\mathcal{M}_{adv}), y^{*}) + \beta \cdot \mathcal{R}(\mathcal{M}_{adv}, \mathcal{M}),\\
\text{where }
  \hat{I}_{\vect{v}}(\mathcal{M}_{adv}) = \mathcal{S}(\mathcal{V}^{*}, \mathcal{T}^{*}, \mathcal{F}, \vect{v}),\nonumber
\label{eq:object}
\end{gather}
where $\hat{I}_{\vect{v}}(\mathcal{M}_{adv})$ represents the rendered image of $\mathcal{M}_{adv}$ under the viewpoint $\vect{v}$ and $V$ is the feasible distribution. $L_{f}$ is the cross-entropy loss~\cite{de2005tutorial} that facilitates the misclassification of $\hat{I}_{\vect{v}}(\mathcal{M}_{adv})$ to $y^{*}$, $R$ is the regularization for minimizing a perceptibility distance between $\mathcal{M}_{adv}$ and $\mathcal{M}$ and $\beta$ is a balancing hyperparameter between these two losses.

To ensure the feasibility of our 3D adversarial example in the real world, we also introduce an EOT scheme (as detailed in \cref{sec:eot}) to help bridge the gap from ``digital" to ``physical". However, the mesh-based optimization by following the objective function~(\ref{eq:object}) needs to calculate gradients in high-dimensional mesh space due to thousands of points in each 3D object. It will easily distort the natural textural appearance of the textured mesh and is trapped into the overfitting with unsatisfactory transferability. 

\subsection{Dual Optimization in NeRF Parameter Space}
\label{sec:dual opt}
In this section, we aim to deviate from the existing mesh-based optimization regime and perform the optimization trajectory in the parameter space of grid-based NeRF~\cite{muller2022instant} as a regularization for escaping from overfitting. 

Specifically, as introduced in \cref{nerf}, grid-based NeRF stores appearance feature vectors in a structured feature grid $G_{tex}(\bm{\cdot};\Theta_{G_{tex}})$ and processes them into emitted colors through a shallow MLP $M_{tex}(\bm{\cdot};\Theta_{M_{tex}})$. Actually, the clean textural parameter set $\mathcal{T}$, representing the colors of all vertices in the initial 3D textured mesh $\mathcal{M}$, are estimated based on the aforementioned mechanism, expressed as follows:
\begin{equation}
\label{eq:colors_map}
\mathcal{T} = \{ \vect{c} | \vect{c} = M_{tex}(\vect{f_{tex}}(\vect{x}); \Theta_{M_{tex}}), \forall \vect{x} \in \mathcal{V} \},
\end{equation}
where $\vect{f_{tex}}(\vect{x})$ is the textual feature vector of $\vect{x}$ and mapped by $G_{tex}(\bm{\cdot};\Theta_{G_{tex}})$ following \cref{map}.

Existing mesh-based optimization methods, which primarily perform direct alterations to the vertex colors in $\mathcal{T}$, struggle with the computing complexity, unsatisfactory transferability, and naturalness inherent in explicit mesh spaces. To prevent this, our method innovatively utilizes the previously established color estimation mechanism, \ie, \cref{eq:colors_map} to perform adversarial fine-tuning in the space of grid-based NeRF, indirectly generating adversarial textures. Moreover, we undertake a dual optimization, concurrently targeting both the parameters of the appearance feature grid, $ \Theta_{G_{tex}} $, and those of the corresponding MLP, $ \Theta_{M_{tex}} $, which effectively integrates the manipulation of detailed textural features contained within the grids with the nuanced processing capabilities inherent in the MLPs.

In essence, such a dual optimization strategy facilitates the embedding of adversarial perturbations at both the foundational feature level and the more sophisticated decision-making echelons of the neural network, thus yielding more transferable 3D adversarial examples while also ensuring naturalness. It is also noteworthy that though merely optimizing the geometric shape proves challenging for achieving transferable targeted 3D attacks, optimizing the texture in conjunction with adjustments to the geometric shape (i\ie, mesh vertex coordinates) can enhance transferability to a certain extent, as verified in \cref{basic}. However, fine-tuning the geometric parameters in the grid-based NeRF space will lead to significant deformation, and ensuring naturalness by restricting the distance to the original mesh's geometry in such a space would require repeatedly using the marching cubes~\cite{lorensen1998marching} technique to convert geometric parameters into mesh during optimization. This incurs a substantial time cost. Therefore, we make a compromise that explicitly modifies the mesh vertex coordinates.

To sum up, we target the grid-based NeRF's inner parameters: $\left(\Theta_{G_{tex}}, \Theta_{M_{tex}}\right)$ instead of vertex colors $\mathcal{T}$ in appearance, accompanying the optimization to mesh's vertex coordinates, thus the objective function~(\ref{eq:object}) becomes:
\begin{gather}
\mathop{\text{min}}\limits_{\mathcal{V}^{*}, \Theta_{G_{tex}}^{*}, \Theta_{M_{tex}^{*}}}\mathbb{E}_{\vect{v}\in V}L_{f}(\hat{I}_{\vect{v}}(\mathcal{M}_{adv}), y^{*}) + 
\\
\beta \cdot \mathcal{R}(\mathcal{M}_{adv}, \mathcal{M}).\nonumber
\label{eq:dual_object}
\end{gather}

\subsection{Regularization for Naturalness}
To ensure the perceptible naturalness of 3D adversarial examples, we impose constraints to both appearance and geometry. For the constraint of appearance, we evaluate the disparity between images rendered under various viewpoints from the adversarial textured mesh $\mathcal{M}_{adv}$ and those rendered from the initial clean textured mesh $\mathcal{M}$, which is achieved by calculating the squared distance between these two sets of images. This appearance-related constraint, denoted as $R_{rgb}$, ensures that the adversarial examples, while effective, do not deviate significantly in appearance from the original mesh and can be formulated as follows:
\begin{equation}
R_{rgb}(\mathcal{M}_{adv}, \mathcal{M}) = \frac{1}{N} \sum_{\vect{v} \in V} \left| \hat{I}_{\vect{v}}(\mathcal{M}_{adv}) - \hat{I}_{\vect{v}}(\mathcal{M}) \right|^{2},
\end{equation}
where $N$ is the sampled number for one epoch.

Then, referring to the constraint in geometry, we incorporate the Chamfer distance $R_{cd}$ to make vertices of $\mathcal{M}_{adv}$ not far from their initial positions in $\mathcal{M}$, the Laplacian Smoothing Loss $R_{lap}$~\cite{nealen2006laplacian} for preventing self-intersecting, and the Mesh Edge Length Loss $R_{edge}$~\cite{wang2018pixel2mesh} for the smoothness of the mesh surface. Above all, the whole regularization $R$ could be represented as follows:
\begin{gather}
    \label{eq:regular}
R(\mathcal{M}_{adv}, \mathcal{M})=\lambda_{1}R_{rgb}(\mathcal{M}_{adv}, \mathcal{M}) + \\ 
\lambda_{2}R_{cd}(\mathcal{M}_{adv}, \mathcal{M}) + \lambda_{3}R_{lap}(\mathcal{M}_{adv}) + \lambda_{4}R_{edge}(\mathcal{M}_{adv}),\nonumber
\end{gather}
where $\lambda_{1}, \lambda_{2}, \lambda_{3}$ and $\lambda_{4}$ are hyperparameters that represent weights of $R_{rgb}$, $R_{cd}$, $R_{lap}$, and $R_{edge}$, respectively. 

\subsection{Physical Attack}
\label{sec:eot}
To realize physically feasible 3D adversarial examples, we need to ensure their robustness against complex transformations in the physical world, including 3D rotations, affine projections, color discrepancies, \etc. A prevalent technique that we utilize here is the Expectation Over Transformation (EOT) algorithm~\cite{athalye2018synthesizing}, which optimizes the adversarial example across a distribution of varying transformations in both 2D and 3D spaces. Specifically, we effectively merge diverse 3D transformations, such as pose, distance, and viewpoint shifts, during rendering with 2D transformations such as contrast and blurring on rendered images. With EOT, the rendered images for optimization in \cref{eq:dual_object} can be as below:
\begin{gather}\label{eq:classify_eot}
  \hat{I}_{\vect{v}}(\mathcal{M}_{adv}) = t(\mathcal{S}(\mathcal{V}^{*}, \mathcal{T}^{*}, \mathcal{F}, \rho(\vect{v}))), \\
  \text{where }
t \in T, \rho \in \mathcal{Q}, \vect{v} \in V,\nonumber
\end{gather}
where $t$ and $\rho$ represent the randomly sampled transformations from $T$ and $\mathcal{Q}$. $T$ and $\mathcal{Q}$ are the sets of various transformations in the 2D space and the 3D space, respectively.

\begin{table*}[!t]
\vspace{-3ex}
\small
\setlength{\tabcolsep}{3pt}
  \centering
  \resizebox{\linewidth}{!}{
  \begin{tabular}{c|c|c|c|c|c|c|c|c|c|c|c|c}
    \hline
   \multirow{2}{*}{ \textbf{Source Model} } & \multirow{2}{*}{ \textbf{Methods} } & \multicolumn{11}{c}{\textbf{Victim Model}} \\ \cline{3-13}
  & &    RN-50 & RN-101 &  RN-152 & VGG-16 & VGG-19 & Inc-v3 &  DN-121 & EN-B0 & MN-v2 & Swin-B & ViT-B/16 \\
    \hline
    \multirow{4}{*}{ResNet-101} 
           & Mesh-based~\cite{xiao2019meshadv}~(enhanced) & 4.91 & \cellcolor{LightCyan}$\bm{99.90^{*}}$ & 8.73  & 1.62 & 0.84 & 4.77  & 3.46 & 2.21  & 1.57 & 2.93  & 2.39  \\\cline{2-13}
   & MLP-only & 40.48 & $56.90^{*}$ & 43.87  & 26.29 & 22.92 & 27.81 & 56.15 & 25.26 & 23.11 & 28.78 & 35.37  \\\cline{2-13}
       & Grid-only &  64.85 & $80.89^{*}$ & 70.49 & 42.89 & 37.29 & 51.60 & 80.59 & 42.54 & 40.24 & 50.07  & 48.66  \\\cline{2-13}
     &  MLP+Grid & \cellcolor{LightCyan}\textbf{78.85} & $88.98^{*}$ &  \cellcolor{LightCyan}\textbf{78.96} & \cellcolor{LightCyan}\textbf{58.29} & \cellcolor{LightCyan}\textbf{51.54} & \cellcolor{LightCyan}\textbf{55.14} & \cellcolor{LightCyan}\textbf{89.14} &  \cellcolor{LightCyan}\textbf{58.71} & \cellcolor{LightCyan}\textbf{57.69} &  \cellcolor{LightCyan}\textbf{69.80} &  \cellcolor{LightCyan}\textbf{55.84} \\\cline{2-13}
    \hline
    \multirow{4}{*}{DenseNet-121} 
               & Mesh-based~\cite{xiao2019meshadv}~(enhanced) & 1.10 & 2.23 & 2.55  & 1.72 & 1.78 & 2.76 & \cellcolor{LightCyan}$\bm{99.12^{*}}$ & 3.91  & 1.39 & 2.74  & 1.68  \\\cline{2-13}
        & MLP-only & 24.58 & 17.68 & 23.18 & 19.30 & 15.69 & 21.68 & $58.04^{*}$ & 17.95 & 13.96 & 17.70 & 22.66 \\\cline{2-13}
               & Grid-only & 45.41 & 31.46 & 39.51 & 30.76 & 29.39 & 37.87 & $89.62^{*}$ & 25.20 & 21.39 & 29.37 & 28.37 \\\cline{2-13}
              & MLP+Grid &  \cellcolor{LightCyan}\textbf{70.98}&  \cellcolor{LightCyan}\textbf{51.61}&  \cellcolor{LightCyan}\textbf{59.59}& \cellcolor{LightCyan}\textbf{57.59} &  \cellcolor{LightCyan}\textbf{50.25}&  \cellcolor{LightCyan}\textbf{43.81} & $96.74^{*}$ &  \cellcolor{LightCyan}\textbf{40.59} &  \cellcolor{LightCyan}\textbf{45.06} & \cellcolor{LightCyan}\textbf{55.15} & \cellcolor{LightCyan}\textbf{36.49} \\
    \hline
  \end{tabular}}
    \caption{The \textbf{ASR(\%)} of 3d adversarial examples generated by different methods including the enhanced mesh-based, the mlp-only, the grid-only, and our dual optimization method under random viewpoints against ResNet-50 (RN-50), ResNet-101 (RN-101), ResNet-152 (RN-152), VGG-16, VGG-19, Inception-v3 (Inc-v3), DenseNet-121 (DN-121), EfficientNet-B0 (EN-B0), MobileNet-v2 (MN-v2), Swin-B, and VIT-B/16. The adversarial examples are learned against the surrogate models ResNet-101 and DenseNet-121.} 
    \label{tab:basic}
\end{table*}
\begin{figure*}
    \centering
    \includegraphics[width=\linewidth]{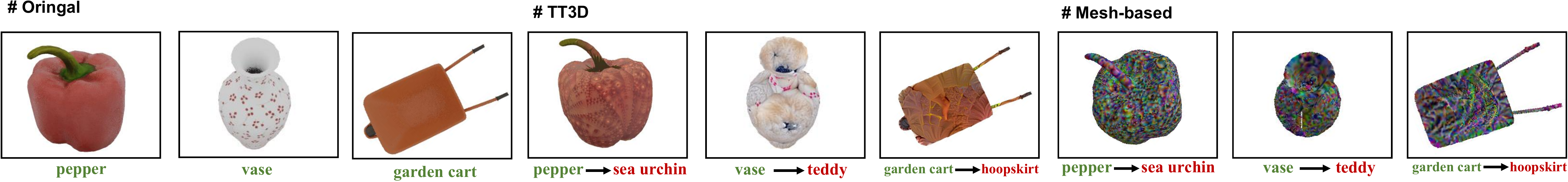}
    \caption{Visual examples of original objects, our TT3D and mesh-based optimization methods under random viewpoints.}
    \label{fig:visual_natural}
    \vspace{-0.3cm}
\end{figure*}
\section{Experiments}
\label{sec:experiments}
In this section, we present the experimental results both in the digital world and the physical world to demonstrate the effectiveness of the proposed method.
\subsection{Experiment Settings}
\label{sec:setting}
\quad\ \textbf{Datasets.} We use the IM3D~\cite{Ruan_2023_ICCV} dataset in our experiments, which contains 1K typical 3D objects from 100 ImageNet~\cite{deng2009imagenet} categories. For our experimental validation, we choose 100 objects randomly from 30 categories.

\textbf{Victim models.} Two typical classifier models, ResNet-101~\cite{He_2016_CVPR} and DenseNet-121~\cite{huang2017densely} are chosen as surrogate models to perform attacks. Other models to test transferability consist of the CNN-based ResNet-50~\cite{He_2016_CVPR}, ResNet-152~\cite{He_2016_CVPR}, VGG-16~\cite{simonyan2014very}, VGG19~\cite{simonyan2014very}, Inceprion-v3~\cite{szegedy2016rethinking}, EfficientNet-B0~\cite{tan2019efficientnet}, MobileNet-V2~\cite{sandler2018mobilenetv2}and the Transformer-based Swin-B~\cite{liu2021swin}, ViT-B/16~\cite{dosovitskiy2020image}. For testing the cross-render transferability, we choose two commercial rendering software: MeshLab and Blender. For testing the cross-task transferability, we choose two other tasks: zero-shot detection~\cite{liu2023grounding} and image caption~\cite{li2022blip} due to their non-limitation to types of objects.

\textbf{Evaluation metrics.} To quantitatively evaluate the effectiveness of our proposed TT3D, we measure the attack success rate (ASR). Specifically, given that our method is a targeted attack, we judge by whether rendered images can be predicted as the target label by victim models. For each 3D object, we render 100 images using random rendering parameters. The ASR for an object is calculated as the proportion of successfully attacked images in 100 rendered images. The final ASR is determined by averaging the ASRs across all reconstructed adversarial objects. 

\textbf{TT3D hyperparameters.} $\beta$ in \cref{eq:dual_object} is $10^{3}$, the epochs are 250, $\lambda_{1}$ is $1$, $\lambda_{2}$ is $3000$, $\lambda_{3}$ is $10^{-3}$, $\lambda_{4}$ is $10^{-2}$.

\subsection{Attack performance in the Digital World}
\subsubsection{Basic Results}
\label{basic}

\begin{table*}[!t]
% \vspace{-3ex}
\small
\setlength{\tabcolsep}{3pt}
  \centering
  \begin{tabular}{c|c|c|c|c|c|c|c|c|c|c|c|c}
    \hline
   \multirow{2}{*}{ \textbf{Source Model} } & \multirow{2}{*}{ \textbf{Methods} } & \multicolumn{11}{c}{\textbf{Victim Model}} \\ \cline{3-13}
  & &    RN-50 & RN-101 &  RN-152 & VGG-16 & VGG-19 & Inc-v3 &  DN-121 & EN-B0 & MN-v2 & Swin-B & ViT-B/16 \\
    \hline
    \multirow{3}{*}{ResNet-101} 
    & Nvdiffrast & \cellcolor{LightCyan}\textbf{78.85} & \cellcolor{LightCyan}$\bm{88.98^{*}}$ &  \cellcolor{LightCyan}\textbf{78.96} & \cellcolor{LightCyan}\textbf{58.29} & \cellcolor{LightCyan}\textbf{51.54} & \cellcolor{LightCyan}\textbf{55.14} & \cellcolor{LightCyan}\textbf{89.14} &  \cellcolor{LightCyan}\textbf{58.71} & \cellcolor{LightCyan}\textbf{57.69} &  \cellcolor{LightCyan}\textbf{69.80} &  \cellcolor{LightCyan}\textbf{55.84} \\\cline{2-13}
   & Meshlab & 53.89 & $60.74^{*}$ & 50.18 & 45.43 & 40.00 & 43.33  & 51.20  & 56.56 & 45.50 & 54.65  & 45.80   \\\cline{2-13}
     &  Blender & 60.03 & $83.88^{*}$ & 51.15  & 39.47 & 40.29 & 49.34  & 60.17 & 36.68 & 43.59   & 50.00  & 48.19  \\\cline{2-13}
    \hline
    \multirow{3}{*}{DenseNet-121} 
        & Nvdiffrast & \cellcolor{LightCyan}\textbf{70.98}&  \cellcolor{LightCyan}\textbf{51.61}&  \cellcolor{LightCyan}\textbf{59.59}& \cellcolor{LightCyan}\textbf{57.59} &  \cellcolor{LightCyan}\textbf{50.25}&  \cellcolor{LightCyan}\textbf{43.81} & \cellcolor{LightCyan}$\bm{96.74^{*}}$ &  \cellcolor{LightCyan}\textbf{40.59} &  \cellcolor{LightCyan}\textbf{45.06} & \cellcolor{LightCyan}\textbf{55.15} & \cellcolor{LightCyan}\textbf{36.49} \\\cline{2-13}
           & Meshlab & 46.34 & 40.33 & 39.43 & 43.24 & 38.00 & 34.00 & $74.67^{*}$ & 43.04 & 29.43 & 44.85  & 33.75 \\\cline{2-13}
              & Blender & 47.66 & 39.06 & 30.94 &  42.34  & 40.00 & 31.25 & $88.75^{*}$ & 29.84 & 26.09 & 39.53 & 26.09 \\
    \hline
  \end{tabular}
    \caption{The \textbf{ASR(\%)} of 3d adversarial examples with different renders against RN-50, RN-101, RN-152, VGG-16, VGG-19, Inc-v3, DN-121, EN-B0, MN-v2, Swin-B, and VIT-B/16. The adversarial examples are learned against surrogate models RN-101 and DN-121.} 
    \label{tab:render}
\end{table*}

\begin{table*}[!t]
\small
\setlength{\tabcolsep}{3pt}
  \centering
  % \resizebox{\linewidth}{!}{
  \begin{tabular}{c|c|c|c|c|c|c|c|c|c|c|c|c}
    \hline
   \multirow{2}{*}{ \textbf{Source Model} } & \multirow{2}{*}{ \textbf{Methods} } & \multicolumn{11}{c}{\textbf{Victim Model}} \\ \cline{3-13}
  & &    RN-50 & RN-101 &  RN-152 & VGG-16 & VGG-19 & Inc-v3 &  DN-121 & EN-B0 & MN-v2 & Swin-B & ViT-B/16 \\
    \hline
    \multirow{2}{*}{ResNet-101} 
    & Tex & 76.96 & \cellcolor{LightCyan}$\bm{91.60^{*}}$ & 78.52 & 52.00 & 49.51 & 53.31 & 72.96 & 55.73 & 53.09  & 64.40 & 54.50 \\\cline{2-13}
     &  Tex+Geo & \cellcolor{LightCyan}\textbf{78.85} & $88.98^{*}$ &  \cellcolor{LightCyan}\textbf{78.96} & \cellcolor{LightCyan}\textbf{58.29} & \cellcolor{LightCyan}\textbf{51.54} & \cellcolor{LightCyan}\textbf{55.14} & \cellcolor{LightCyan}\textbf{89.14} &  \cellcolor{LightCyan}\textbf{58.71} & \cellcolor{LightCyan}\textbf{57.69} &  \cellcolor{LightCyan}\textbf{69.80} &  \cellcolor{LightCyan}\textbf{55.84} \\\cline{2-13}
    \hline
    \multirow{2}{*}{DenseNet-121} 
        & Tex & 65.71 & 46.13 & 52.13 & 46.60 & 40.85 & 36.44 & $94.28^{*}$ & 37.49 & 38.72 & 47.62 & 35.58 \\\cline{2-13}
              & Tex+Geo &  \cellcolor{LightCyan}\textbf{70.98}&  \cellcolor{LightCyan}\textbf{51.61}&  \cellcolor{LightCyan}\textbf{59.59}& \cellcolor{LightCyan}\textbf{57.59} &  \cellcolor{LightCyan}\textbf{50.25}&  \cellcolor{LightCyan}\textbf{43.81} & \cellcolor{LightCyan}$\bm{96.74^{*}}$ &  \cellcolor{LightCyan}\textbf{40.59} &  \cellcolor{LightCyan}\textbf{45.06} & \cellcolor{LightCyan}\textbf{55.15} & \cellcolor{LightCyan}\textbf{36.49} \\
    \hline
  \end{tabular}
    \caption{The \textbf{ASR(\%)} of tex-only and tex+geo(sub) optimization methods against RN-50, RN-101, RN-152, VGG-16, VGG-19, Inc-v3, DN-121, EN-B0, MN-v2, Swin-B, and VIT-B/16. The adversarial examples are learned against the surrogate models RN-101 and DN-121.} 
    \label{tab:tex+geo}
    \vspace{-2ex}
\end{table*}

\quad\ \textbf{Effectiveness of the proposed method.} 
To verify the effects of our proposed method, we compare the attack performance with different methods. Among them, for a fair comparison with mesh-based optimization method~\cite{xiao2019meshadv}, we enhance it to optimize both the vertex colors and the coordinates. Specifically, this method is based on the classical method MeshAdv~\cite{xiao2019meshadv} and to further accentuate the limitations of this method, we put no limitations on the color modifications.
 \cref{tab:basic} shows the attack success rates ($\%$) in various common classifier models, where we find that mesh-based optimization exhibits nearly no transferability due to its tendency to overfit. In contrast, the dual optimization targetting both MLP and Grid parameters demonstrates a substantial improvement in transferability. Then, to explore the effects of different parameters in grid-based NeRF space, we also test the attack performance of MLP-only optimization and Grid-only optimization. Experimental results show that MLP-only optimization shows reasonable success, indicating its effectiveness in manipulating model-specific features. However, Grid-only optimization presents better transferability, likely due to its ability to capture spatial relationships more effectively. The combination of both, as evidenced by the highest success rates across victim models in \cref{tab:basic}, suggests that this dual optimization method effectively integrates detailed textural features and complex decision-making processes, proving the necessity and superiority of the dual optimization strategy.

\textbf{Better visual naturalness.}
In TT3D, while achieving excellent attack effectiveness, our method also exhibits visual naturalness compared to mesh-based optimization methods. As illustrated in \cref{fig:visual_natural}, 3D objects optimized using mesh-based methods display non-sensical noise for appearance. In contrast, our TT3D, benefiting from targeting on both the foundation feature level and decision layers, could generate adversarial perturbations that are more semantically informative and show greater resemblance to the original ones. More specific details, including quantitative metrics, are provided in the \textit{Supplementary Materials}.
% \vspace{-1ex}

\subsubsection{Cross-render Transferability}
\begin{figure}[!h]
    \centering
    \includegraphics[width=0.9\linewidth]{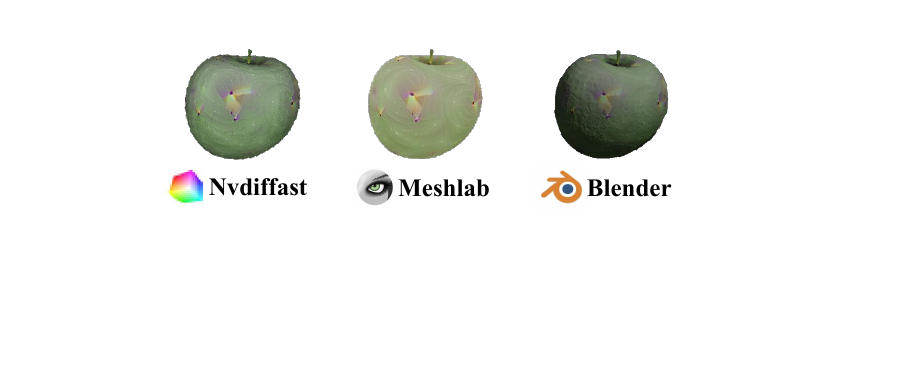}
        \vspace{-0.4cm}
    \caption{Comparison between rendering results of different renders, including differential rendering library Nvdiffast, commercial software Meshlab, and Blender.}
    \label{fig:cross-render}
\end{figure}

As illustrated in \cref{fig:cross-render}, we observe that there exist variations in the rendering results of different renderers. To further validate the effectiveness of our method, we conduct transferability tests across various renderers, including two commercial-grade rendering software. As shown in \cref{tab:render}, our 3D adversarial examples still perform well, even when faced with completely unknown rendering systems. This demonstrates the robustness of our method, as there is no significant performance degradation across different rendering environments, highlighting the broad applicability and resilience of our adversarial examples.

\subsubsection{Cross-task Transferability}
To verify the transferability of our adversarial examples across different tasks, we select zero-shot detection and image caption for testing due to their broad applicability.

\begin{figure}[!h]
    \centering
    \includegraphics[width=\linewidth]{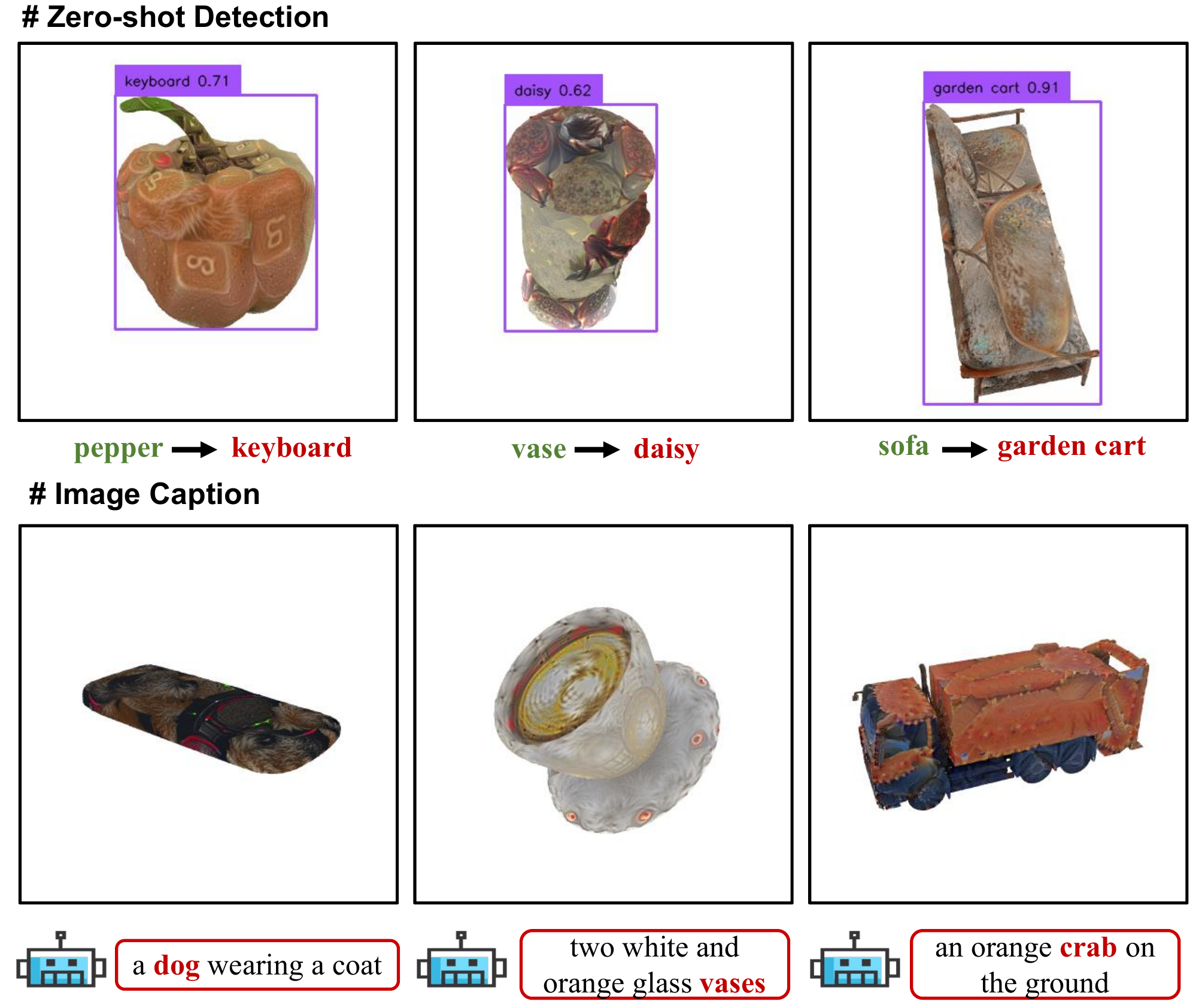}
    \caption{The prediction examples of our 3D adversarial examples targeting zero-shot detection and image caption tasks. Green
and red text represent the clean label and the target label, respectively.}
\vspace{-2ex}
    \label{fig:cross-task}

\end{figure}

\textbf{Zero-shot detector.} 
We utilize the state-of-the-art model of Liu \etal~\cite{liu2023grounding} for zero-shot detection. In our tests, we eliminate background distractions by using a plain white background, focusing solely on the detector's ability to identify the target label in the rendered adversarial sample images. A detection threshold greater than 0.5 is considered successful. Experimental results reveal that even in this setting, our method achieved a success rate of 76.94\%, as evidenced by some successful examples shown in \cref{fig:cross-task}.

\textbf{Image caption.}
For image captioning, we employ the BLIP model from Li et al.~\cite{li2022blip}, still maintaining the white background setup. We present the rendered adversarial images to the BLIP model and deem the test successful if the generated caption includes the target label. Due to testing costs, we randomly selected three viewpoints for 100 different 3D objects. The results demonstrate that our method still achieves a success rate of 32.33\% in this task, as evidenced by some successful examples shown in \cref{fig:cross-task}.

\begin{figure*}
    \centering
    \includegraphics[width=0.95\linewidth]{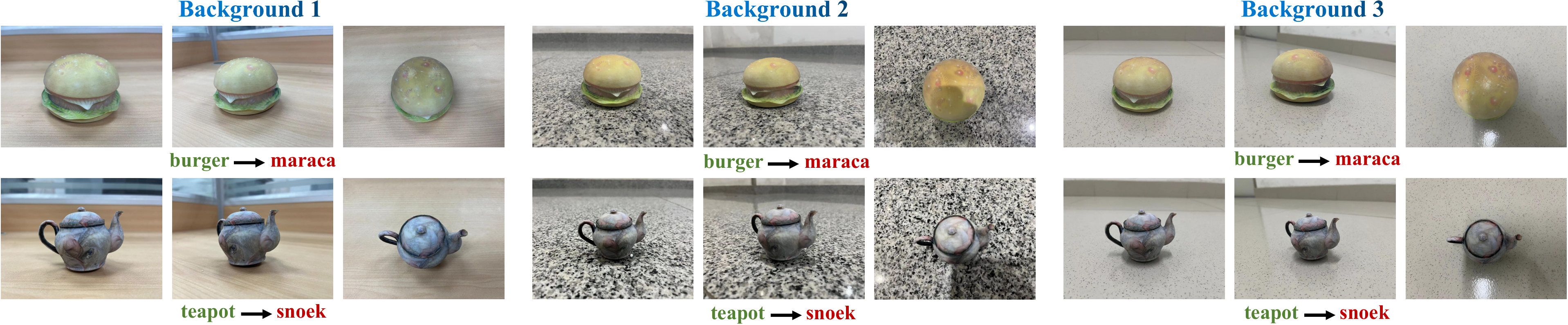}
    \caption{Visual examples of printed 3D adversarial objects towards the target label under different backgrounds ({{\color{blue}B-1}},{{\color{blue}B-2}}, {{\color{blue}B-3}}) and viewpoints in the physical world. The first row is learned against ResNet-101 and the second row is learned against DenseNet-121.}
    \label{fig:physical}
    \vspace{-1ex}
\end{figure*}

\subsection{Additional Results and Ablation Study}
\quad\ \textbf{The effects of additional vertex optimization.}
To validate the additional vertex optimization mentioned in \cref{sec:problem} that could serve as an auxiliary means to further enhance texture-focused optimization performance, we conduct an ablation study on it. As shown in \cref{tab:tex+geo}, it is evident that adding the geometric changes of the mesh vertex coordinates while optimizing texture features in the grid-based parameter space can, to a certain extent, enhance the transferability of the attack. Additionally, we find that solely altering the geometric structure with requirements for naturalness, hardly achieves successful targeted attacks with arbitrary objectives. 
Therefore, we have not included results for solely optimization geometric structures in \cref{tab:tex+geo}.
\begin{figure}
    \centering
    \includegraphics[width=0.9\linewidth]{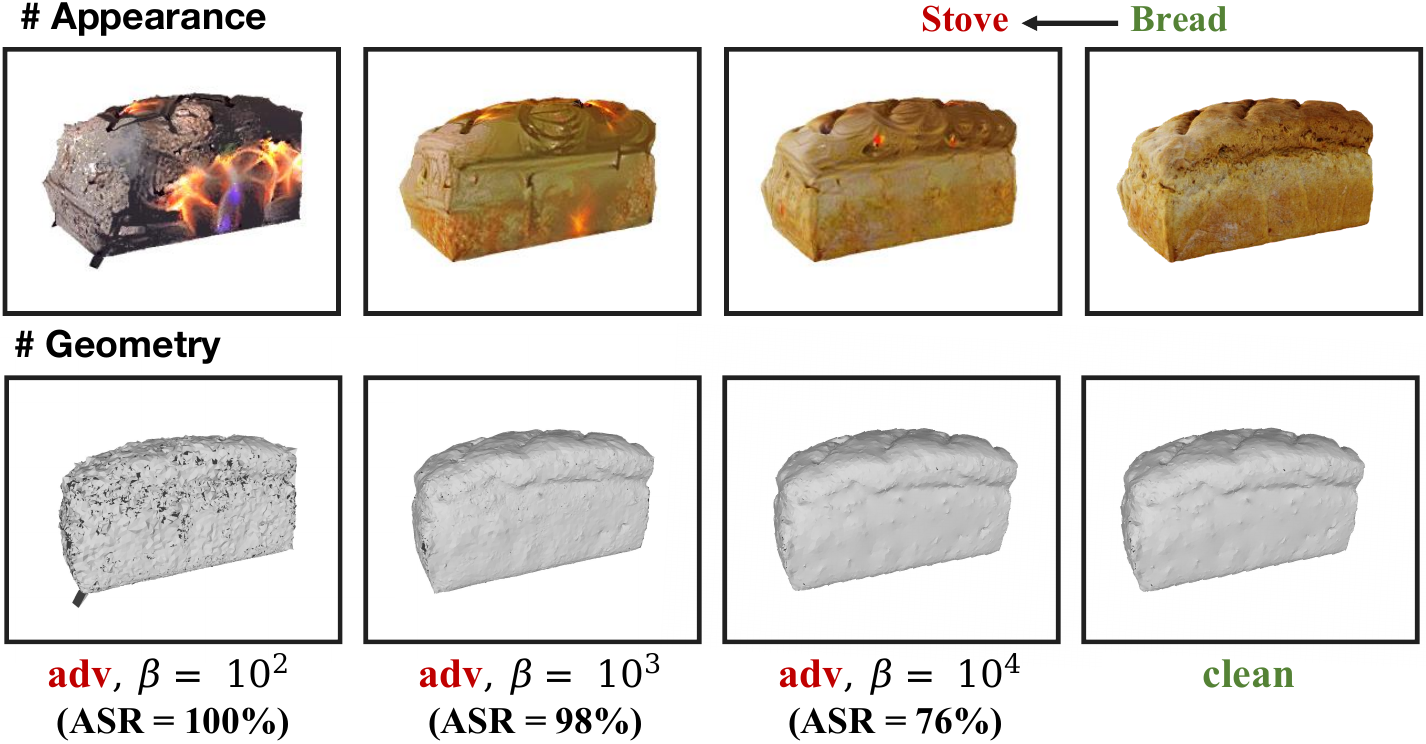}
    \caption{An visual example of the change in appearance and geometry with $\beta$ and its corresponding attack success rate.}
    \label{fig:beta}
    \vspace{-1ex}
\end{figure}

\textbf{The effects of $\beta$ in $\mathcal{R}$.}
$\beta$ as a hyperparameter to adjust the regularization $
\mathcal{R}$ responsible for naturalness, has a significant impact on both naturalness and attack performance. Thus, to fully explore its impacts, we conduct an ablation study on the variation of $\beta$. As illustrated in \cref{fig:beta}, when $\beta$ increases from lower to higher values (from $10^{2}$, $10^{3}$ to $10^{4}$), the distance from both the appearance and geometric structure of the adversarial sample to the original 3D object becomes closer, making the perturbation more concealed, but accompanying an observed decline in the overall success rate of the attack (detailed in the \textit{ Supplementary Materials}). 
To strike a balance between attack performance and naturalness, we ultimately choose a $\beta$ value of $10^{3}$. It is also noteworthy that, regardless of the parameter setting, our adversarial examples feature textures not in the form of noise, but as semantically meaningful textures. This approach is more natural compared to explicit optimization in mesh-based optimization methods, as shown in \cref{fig:visual_natural}.

\subsection{Attack Performance in the Physical World}
\label{sec:phy}
To verify the feasibility of TT3D in the physical world, we first utilize 3D printing techniques to print 3D adversarial objects. Then, considering that the backgrounds in the physical world are diverse, unlike the white background of the optimization process, we not only validate the digital-to-physical transferability of TT3D under different viewpoints, but also add three randomly selected, distinct backgrounds {{\color{blue}B-1}},{{\color{blue}B-2}}, and {{\color{blue}B-3}} to evaluate its robustness against backgrounds. For the number of testing objects, we print 20 3d objects, including 10 for the surrogate ResNet-101 and 10 for the surrogate DenseNet-121. The specific process involves 1) placing the 3D adversarial objects on the surface; 2) slowly circling them with a smartphone for about $360^{\circ}$, excluding the bottom part. which lasts approximately 20 seconds per object in each setting, capturing 10 frames per second, resulting in a total of 200 frames; 3) calculating the attack success rate, which is determined by the proportion of successful frames. The experimental results are presented in \cref{fig:physical} and \cref{tab:sys}, demonstrating the robust effectiveness of our method in various scenarios from the physical world. Further experiments on transferability tests in the physical world are available in \textit{Supplementary Materials}.

\begin{table}[!h]
    \begin{center}
    % \small %footnotesize
    \setlength{\tabcolsep}{7pt}
    \begin{tabular}{c|ccc}
    \hline 
        & {{\color{blue}B-1}} & {{\color{blue}B-2}} & {{\color{blue}B-3}} \\
        \hline
      ResNet-101 & 81.30 & 76.45 & 91.55 \\
       \hline      
       DenseNet-121 & 79.30  & 83.25 & 84.50 \\
       \hline
    \end{tabular}
            \caption{The ASR(\%) of printed adversarial meshes against the RN-101 and DN-121 under different backgrounds: {\color{blue}B-1}, {\color{blue}B-2}, {\color{blue}B-3} and various viewpoints in the real wolrd.}
    \label{tab:sys}
    \vspace{-3ex}
    \end{center}

\end{table}
\section{Conclusion}
\label{sec:conclusion}
In this paper, we propose a novel 3D attack framework \textbf{TT3D} that could rapidly reconstruct multi-view images into transferable targeted 3D adversarial examples, effectively filling the gap in 3D transferable targeted attacks. Besides, TT3D leverages a dual optimization strategy in the grid-based NeRF space, which significantly improves the black-box transferability meanwhile ensuring visual naturalness. Extensive experiments further demonstrate TT3D's strong cross-model transferability and adaptability in various renders and vision tasks, with real-world applicability confirmed through 3D printing techniques.
\section*{Acknowledgement}
This work was supported in part by the Project of the National
Natural Science Foundation of China (Nos. 62076018, U2341228, 62276149) and in part by the
Fundamental Research Funds for the Central Universities.
{
    \small
    \bibliographystyle{ieeenat_fullname}
    \bibliography{main}

\begin{thebibliography}{57}
\providecommand{\natexlab}[1]{#1}
\providecommand{\url}[1]{\texttt{#1}}
\expandafter\ifx\csname urlstyle\endcsname\relax
  \providecommand{\doi}[1]{doi: #1}\else
  \providecommand{\doi}{doi: \begingroup \urlstyle{rm}\Url}\fi

\bibitem[Athalye et~al.(2018)Athalye, Engstrom, Ilyas, and Kwok]{athalye2018synthesizing}
Anish Athalye, Logan Engstrom, Andrew Ilyas, and Kevin Kwok.
\newblock Synthesizing robust adversarial examples.
\newblock In \emph{International conference on machine learning}, pages 284--293. PMLR, 2018.

\bibitem[De~Boer et~al.(2005)De~Boer, Kroese, Mannor, and Rubinstein]{de2005tutorial}
Pieter-Tjerk De~Boer, Dirk~P Kroese, Shie Mannor, and Reuven~Y Rubinstein.
\newblock A tutorial on the cross-entropy method.
\newblock \emph{Annals of operations research}, 134:\penalty0 19--67, 2005.

\bibitem[Deng et~al.(2009)Deng, Dong, Socher, Li, Li, and Fei-Fei]{deng2009imagenet}
Jia Deng, Wei Dong, Richard Socher, Li-Jia Li, Kai Li, and Li Fei-Fei.
\newblock Imagenet: A large-scale hierarchical image database.
\newblock In \emph{2009 IEEE conference on computer vision and pattern recognition}, pages 248--255. Ieee, 2009.

\bibitem[Dong et~al.(2018)Dong, Liao, Pang, Su, Zhu, Hu, and Li]{dong2018boosting}
Yinpeng Dong, Fangzhou Liao, Tianyu Pang, Hang Su, Jun Zhu, Xiaolin Hu, and Jianguo Li.
\newblock Boosting adversarial attacks with momentum.
\newblock In \emph{Proceedings of the IEEE conference on computer vision and pattern recognition}, pages 9185--9193, 2018.

\bibitem[Dong et~al.(2022{\natexlab{a}})Dong, Ruan, Su, Kang, Wei, and Zhu]{dong2022viewfool}
Yinpeng Dong, Shouwei Ruan, Hang Su, Caixin Kang, Xingxing Wei, and Jun Zhu.
\newblock Viewfool: Evaluating the robustness of visual recognition to adversarial viewpoints.
\newblock \emph{Advances in Neural Information Processing Systems}, 35:\penalty0 36789--36803, 2022{\natexlab{a}}.

\bibitem[Dong et~al.(2022{\natexlab{b}})Dong, Zhu, Gao, et~al.]{dong2022isometric}
Yinpeng Dong, Jun Zhu, Xiao-Shan Gao, et~al.
\newblock Isometric 3d adversarial examples in the physical world.
\newblock \emph{Advances in Neural Information Processing Systems}, 35:\penalty0 19716--19731, 2022{\natexlab{b}}.

\bibitem[Dosovitskiy et~al.(2020)Dosovitskiy, Beyer, Kolesnikov, Weissenborn, Zhai, Unterthiner, Dehghani, Minderer, Heigold, Gelly, et~al.]{dosovitskiy2020image}
Alexey Dosovitskiy, Lucas Beyer, Alexander Kolesnikov, Dirk Weissenborn, Xiaohua Zhai, Thomas Unterthiner, Mostafa Dehghani, Matthias Minderer, Georg Heigold, Sylvain Gelly, et~al.
\newblock An image is worth 16x16 words: Transformers for image recognition at scale.
\newblock \emph{arXiv preprint arXiv:2010.11929}, 2020.

\bibitem[Fridovich-Keil et~al.(2022)Fridovich-Keil, Yu, Tancik, Chen, Recht, and Kanazawa]{fridovich2022plenoxels}
Sara Fridovich-Keil, Alex Yu, Matthew Tancik, Qinhong Chen, Benjamin Recht, and Angjoo Kanazawa.
\newblock Plenoxels: Radiance fields without neural networks.
\newblock In \emph{Proceedings of the IEEE/CVF Conference on Computer Vision and Pattern Recognition}, pages 5501--5510, 2022.

\bibitem[Goodfellow et~al.(2014)Goodfellow, Shlens, and Szegedy]{goodfellow2014explaining}
Ian~J Goodfellow, Jonathon Shlens, and Christian Szegedy.
\newblock Explaining and harnessing adversarial examples.
\newblock \emph{arXiv preprint arXiv:1412.6572}, 2014.

\bibitem[He et~al.(2016)He, Zhang, Ren, and Sun]{He_2016_CVPR}
Kaiming He, Xiangyu Zhang, Shaoqing Ren, and Jian Sun.
\newblock Deep residual learning for image recognition.
\newblock In \emph{Proceedings of the IEEE Conference on Computer Vision and Pattern Recognition (CVPR)}, 2016.

\bibitem[Huang et~al.(2017)Huang, Liu, Van Der~Maaten, and Weinberger]{huang2017densely}
Gao Huang, Zhuang Liu, Laurens Van Der~Maaten, and Kilian~Q Weinberger.
\newblock Densely connected convolutional networks.
\newblock In \emph{Proceedings of the IEEE conference on computer vision and pattern recognition}, pages 4700--4708, 2017.

\bibitem[Laine et~al.(2020)Laine, Hellsten, Karras, Seol, Lehtinen, and Aila]{Laine2020diffrast}
Samuli Laine, Janne Hellsten, Tero Karras, Yeongho Seol, Jaakko Lehtinen, and Timo Aila.
\newblock Modular primitives for high-performance differentiable rendering.
\newblock \emph{ACM Transactions on Graphics}, 39\penalty0 (6), 2020.

\bibitem[Li et~al.(2021)Li, Xie, Wei, and Wenting]{bo2021ship}
Bo Li, Xiaoyang Xie, Xingxing Wei, and Tang Wenting.
\newblock Ship detection and classification from optical remote sensing images: A survey.
\newblock \emph{Chinese Journal of Aeronautics}, 34\penalty0 (3):\penalty0 145--163, 2021.

\bibitem[Li et~al.(2022)Li, Li, Xiong, and Hoi]{li2022blip}
Junnan Li, Dongxu Li, Caiming Xiong, and Steven Hoi.
\newblock Blip: Bootstrapping language-image pre-training for unified vision-language understanding and generation.
\newblock In \emph{International Conference on Machine Learning}, pages 12888--12900. PMLR, 2022.

\bibitem[Li et~al.(2020)Li, Deng, Li, Yan, Gao, and Huang]{li2020towards}
Maosen Li, Cheng Deng, Tengjiao Li, Junchi Yan, Xinbo Gao, and Heng Huang.
\newblock Towards transferable targeted attack.
\newblock In \emph{Proceedings of the IEEE/CVF Conference on Computer Vision and Pattern Recognition}, pages 641--649, 2020.

\bibitem[Liu et~al.(2023)Liu, Zeng, Ren, Li, Zhang, Yang, Li, Yang, Su, Zhu, et~al.]{liu2023grounding}
Shilong Liu, Zhaoyang Zeng, Tianhe Ren, Feng Li, Hao Zhang, Jie Yang, Chunyuan Li, Jianwei Yang, Hang Su, Jun Zhu, et~al.
\newblock Grounding dino: Marrying dino with grounded pre-training for open-set object detection.
\newblock \emph{arXiv preprint arXiv:2303.05499}, 2023.

\bibitem[Liu et~al.(2017)Liu, Chen, Liu, and Song]{Liu2016}
Yanpei Liu, Xinyun Chen, Chang Liu, and Dawn Song.
\newblock Delving into transferable adversarial examples and black-box attacks.
\newblock In \emph{International Conference on Learning Representations}, 2017.

\bibitem[Liu et~al.(2021)Liu, Lin, Cao, Hu, Wei, Zhang, Lin, and Guo]{liu2021swin}
Ze Liu, Yutong Lin, Yue Cao, Han Hu, Yixuan Wei, Zheng Zhang, Stephen Lin, and Baining Guo.
\newblock Swin transformer: Hierarchical vision transformer using shifted windows.
\newblock In \emph{Proceedings of the IEEE/CVF international conference on computer vision}, pages 10012--10022, 2021.

\bibitem[Lorensen and Cline(1998)]{lorensen1998marching}
William~E Lorensen and Harvey~E Cline.
\newblock Marching cubes: A high resolution 3d surface construction algorithm.
\newblock In \emph{Seminal graphics: pioneering efforts that shaped the field}, pages 347--353. 1998.

\bibitem[Madry et~al.(2017)Madry, Makelov, Schmidt, Tsipras, and Vladu]{madry2017towards}
Aleksander Madry, Aleksandar Makelov, Ludwig Schmidt, Dimitris Tsipras, and Adrian Vladu.
\newblock Towards deep learning models resistant to adversarial attacks.
\newblock \emph{arXiv preprint arXiv:1706.06083}, 2017.

\bibitem[Martin-Brualla et~al.(2021)Martin-Brualla, Radwan, Sajjadi, Barron, Dosovitskiy, and Duckworth]{martin2021nerf}
Ricardo Martin-Brualla, Noha Radwan, Mehdi~SM Sajjadi, Jonathan~T Barron, Alexey Dosovitskiy, and Daniel Duckworth.
\newblock Nerf in the wild: Neural radiance fields for unconstrained photo collections.
\newblock In \emph{Proceedings of the IEEE/CVF Conference on Computer Vision and Pattern Recognition}, pages 7210--7219, 2021.

\bibitem[Moosavi-Dezfooli et~al.(2016)Moosavi-Dezfooli, Fawzi, and Frossard]{moosavi2016deepfool}
Seyed-Mohsen Moosavi-Dezfooli, Alhussein Fawzi, and Pascal Frossard.
\newblock Deepfool: a simple and accurate method to fool deep neural networks.
\newblock In \emph{Proceedings of the IEEE conference on computer vision and pattern recognition}, pages 2574--2582, 2016.

\bibitem[M{\"u}ller et~al.(2022)M{\"u}ller, Evans, Schied, and Keller]{muller2022instant}
Thomas M{\"u}ller, Alex Evans, Christoph Schied, and Alexander Keller.
\newblock Instant neural graphics primitives with a multiresolution hash encoding.
\newblock \emph{ACM Transactions on Graphics (ToG)}, 41\penalty0 (4):\penalty0 1--15, 2022.

\bibitem[Naseer et~al.(2021)Naseer, Khan, Hayat, Khan, and Porikli]{naseer2021generating}
Muzammal Naseer, Salman Khan, Munawar Hayat, Fahad~Shahbaz Khan, and Fatih Porikli.
\newblock On generating transferable targeted perturbations.
\newblock In \emph{Proceedings of the IEEE/CVF International Conference on Computer Vision}, pages 7708--7717, 2021.

\bibitem[Nealen et~al.(2006)Nealen, Igarashi, Sorkine, and Alexa]{nealen2006laplacian}
Andrew Nealen, Takeo Igarashi, Olga Sorkine, and Marc Alexa.
\newblock Laplacian mesh optimization.
\newblock In \emph{Proceedings of the 4th international conference on Computer graphics and interactive techniques in Australasia and Southeast Asia}, pages 381--389, 2006.

\bibitem[Oslund et~al.(2022)Oslund, Washington, So, Chen, and Ji]{oslund2022multiview}
Scott Oslund, Clayton Washington, Andrew So, Tingting Chen, and Hao Ji.
\newblock Multiview robust adversarial stickers for arbitrary objects in the physical world.
\newblock \emph{Journal of Computational and Cognitive Engineering}, 1\penalty0 (4):\penalty0 152--158, 2022.

\bibitem[Rawat and Wang(2017)]{rawat2017deep}
Waseem Rawat and Zenghui Wang.
\newblock Deep convolutional neural networks for image classification: A comprehensive review.
\newblock \emph{Neural computation}, 29\penalty0 (9):\penalty0 2352--2449, 2017.

\bibitem[Ruan et~al.(2023{\natexlab{a}})Ruan, Dong, Su, Peng, Chen, and Wei]{Ruan_2023_ICCV}
Shouwei Ruan, Yinpeng Dong, Hang Su, Jianteng Peng, Ning Chen, and Xingxing Wei.
\newblock Towards viewpoint-invariant visual recognition via adversarial training.
\newblock In \emph{Proceedings of the IEEE/CVF International Conference on Computer Vision (ICCV)}, pages 4709--4719, 2023{\natexlab{a}}.

\bibitem[Ruan et~al.(2023{\natexlab{b}})Ruan, Dong, Su, Peng, Chen, and Wei]{ruan2023improving}
Shouwei Ruan, Yinpeng Dong, Hang Su, Jianteng Peng, Ning Chen, and Xingxing Wei.
\newblock Improving viewpoint robustness for visual recognition via adversarial training.
\newblock \emph{arXiv preprint arXiv:2307.11528}, 2023{\natexlab{b}}.

\bibitem[Sandler et~al.(2018)Sandler, Howard, Zhu, Zhmoginov, and Chen]{sandler2018mobilenetv2}
Mark Sandler, Andrew Howard, Menglong Zhu, Andrey Zhmoginov, and Liang-Chieh Chen.
\newblock Mobilenetv2: Inverted residuals and linear bottlenecks.
\newblock In \emph{Proceedings of the IEEE conference on computer vision and pattern recognition}, pages 4510--4520, 2018.

\bibitem[Simonyan and Zisserman(2014)]{simonyan2014very}
Karen Simonyan and Andrew Zisserman.
\newblock Very deep convolutional networks for large-scale image recognition.
\newblock \emph{arXiv preprint arXiv:1409.1556}, 2014.

\bibitem[Sun et~al.(2022)Sun, Sun, and Chen]{sun2022direct}
Cheng Sun, Min Sun, and Hwann-Tzong Chen.
\newblock Direct voxel grid optimization: Super-fast convergence for radiance fields reconstruction.
\newblock In \emph{Proceedings of the IEEE/CVF Conference on Computer Vision and Pattern Recognition}, pages 5459--5469, 2022.

\bibitem[Suryanto et~al.(2022)Suryanto, Kim, Kang, Larasati, Yun, Le, Yang, Oh, and Kim]{suryanto2022dta}
Naufal Suryanto, Yongsu Kim, Hyoeun Kang, Harashta~Tatimma Larasati, Youngyeo Yun, Thi-Thu-Huong Le, Hunmin Yang, Se-Yoon Oh, and Howon Kim.
\newblock Dta: Physical camouflage attacks using differentiable transformation network.
\newblock In \emph{Proceedings of the IEEE/CVF Conference on Computer Vision and Pattern Recognition}, pages 15305--15314, 2022.

\bibitem[Suryanto et~al.(2023)Suryanto, Kim, Larasati, Kang, Le, Hong, Yang, Oh, and Kim]{suryanto2023active}
Naufal Suryanto, Yongsu Kim, Harashta~Tatimma Larasati, Hyoeun Kang, Thi-Thu-Huong Le, Yoonyoung Hong, Hunmin Yang, Se-Yoon Oh, and Howon Kim.
\newblock Active: Towards highly transferable 3d physical camouflage for universal and robust vehicle evasion.
\newblock In \emph{Proceedings of the IEEE/CVF International Conference on Computer Vision}, pages 4305--4314, 2023.

\bibitem[Szegedy et~al.(2013)Szegedy, Zaremba, Sutskever, Bruna, Erhan, Goodfellow, and Fergus]{szegedy2013intriguing}
Christian Szegedy, Wojciech Zaremba, Ilya Sutskever, Joan Bruna, Dumitru Erhan, Ian Goodfellow, and Rob Fergus.
\newblock Intriguing properties of neural networks.
\newblock \emph{arXiv preprint arXiv:1312.6199}, 2013.

\bibitem[Szegedy et~al.(2016)Szegedy, Vanhoucke, Ioffe, Shlens, and Wojna]{szegedy2016rethinking}
Christian Szegedy, Vincent Vanhoucke, Sergey Ioffe, Jon Shlens, and Zbigniew Wojna.
\newblock Rethinking the inception architecture for computer vision.
\newblock In \emph{Proceedings of the IEEE conference on computer vision and pattern recognition}, pages 2818--2826, 2016.

\bibitem[Tan and Le(2019)]{tan2019efficientnet}
Mingxing Tan and Quoc Le.
\newblock Efficientnet: Rethinking model scaling for convolutional neural networks.
\newblock In \emph{International conference on machine learning}, pages 6105--6114. PMLR, 2019.

\bibitem[Tang et~al.(2023)Tang, Zhou, Chen, Hu, Ding, Wang, and Zeng]{tang2023delicate}
Jiaxiang Tang, Hang Zhou, Xiaokang Chen, Tianshu Hu, Errui Ding, Jingdong Wang, and Gang Zeng.
\newblock Delicate textured mesh recovery from nerf via adaptive surface refinement.
\newblock \emph{arXiv preprint arXiv:2303.02091}, 2023.

\bibitem[Thys et~al.(2019)Thys, Van~Ranst, and Goedem{\'e}]{thys2019fooling}
Simen Thys, Wiebe Van~Ranst, and Toon Goedem{\'e}.
\newblock Fooling automated surveillance cameras: adversarial patches to attack person detection.
\newblock In \emph{Proceedings of the IEEE/CVF conference on computer vision and pattern recognition workshops}, pages 0--0, 2019.

\bibitem[Wang et~al.(2022)Wang, Jiang, Sun, Zhou, Gong, Zhang, Yao, and Chen]{wang2022fca}
Donghua Wang, Tingsong Jiang, Jialiang Sun, Weien Zhou, Zhiqiang Gong, Xiaoya Zhang, Wen Yao, and Xiaoqian Chen.
\newblock Fca: Learning a 3d full-coverage vehicle camouflage for multi-view physical adversarial attack.
\newblock In \emph{Proceedings of the AAAI conference on artificial intelligence}, pages 2414--2422, 2022.

\bibitem[Wang et~al.(2018)Wang, Zhang, Li, Fu, Liu, and Jiang]{wang2018pixel2mesh}
Nanyang Wang, Yinda Zhang, Zhuwen Li, Yanwei Fu, Wei Liu, and Yu-Gang Jiang.
\newblock Pixel2mesh: Generating 3d mesh models from single rgb images.
\newblock In \emph{Proceedings of the European conference on computer vision (ECCV)}, pages 52--67, 2018.

\bibitem[Wang et~al.(2004)Wang, Bovik, Sheikh, and Simoncelli]{wang2004image}
Zhou Wang, Alan~C Bovik, Hamid~R Sheikh, and Eero~P Simoncelli.
\newblock Image quality assessment: from error visibility to structural similarity.
\newblock \emph{IEEE transactions on image processing}, 13\penalty0 (4):\penalty0 600--612, 2004.

\bibitem[Wang et~al.(2023)Wang, Yang, Feng, Sun, Guo, Zhang, and Ren]{wang2023towards}
Zhibo Wang, Hongshan Yang, Yunhe Feng, Peng Sun, Hengchang Guo, Zhifei Zhang, and Kui Ren.
\newblock Towards transferable targeted adversarial examples.
\newblock In \emph{Proceedings of the IEEE/CVF Conference on Computer Vision and Pattern Recognition}, pages 20534--20543, 2023.

\bibitem[Wei et~al.(2023{\natexlab{a}})Wei, Guo, and Yu]{9779913}
Xingxing Wei, Ying Guo, and Jie Yu.
\newblock Adversarial sticker: A stealthy attack method in the physical world.
\newblock \emph{IEEE Transactions on Pattern Analysis and Machine Intelligence}, 45\penalty0 (3):\penalty0 2711--2725, 2023{\natexlab{a}}.

\bibitem[Wei et~al.(2023{\natexlab{b}})Wei, Guo, Yu, and Zhang]{9999043}
Xingxing Wei, Ying Guo, Jie Yu, and Bo Zhang.
\newblock Simultaneously optimizing perturbations and positions for black-box adversarial patch attacks.
\newblock \emph{IEEE Transactions on Pattern Analysis and Machine Intelligence}, 45\penalty0 (7):\penalty0 9041--9054, 2023{\natexlab{b}}.

\bibitem[Wei et~al.(2023{\natexlab{c}})Wei, Huang, Sun, and Yu]{Wei_2023_ICCV}
Xingxing Wei, Yao Huang, Yitong Sun, and Jie Yu.
\newblock Unified adversarial patch for cross-modal attacks in the physical world.
\newblock In \emph{Proceedings of the IEEE/CVF International Conference on Computer Vision (ICCV)}, pages 4445--4454, 2023{\natexlab{c}}.

\bibitem[Wei et~al.(2023{\natexlab{d}})Wei, Huang, Sun, and Yu]{wei2023unified}
Xingxing Wei, Yao Huang, Yitong Sun, and Jie Yu.
\newblock Unified adversarial patch for visible-infrared cross-modal attacks in the physical world.
\newblock \emph{IEEE Transactions on Pattern Analysis and Machine Intelligence}, 2023{\natexlab{d}}.

\bibitem[Wei et~al.(2023{\natexlab{e}})Wei, Yu, and Huang]{Wei_2023_CVPR}
Xingxing Wei, Jie Yu, and Yao Huang.
\newblock Physically adversarial infrared patches with learnable shapes and locations.
\newblock In \emph{Proceedings of the IEEE/CVF Conference on Computer Vision and Pattern Recognition (CVPR)}, pages 12334--12342, 2023{\natexlab{e}}.

\bibitem[Wei et~al.(2023{\natexlab{f}})Wei, Yu, and Huang]{wei2023infrared}
Xingxing Wei, Jie Yu, and Yao Huang.
\newblock Infrared adversarial patches with learnable shapes and locations in the physical world.
\newblock \emph{International Journal of Computer Vision}, pages 1--17, 2023{\natexlab{f}}.

\bibitem[Xiao et~al.(2019)Xiao, Yang, Li, Deng, and Liu]{xiao2019meshadv}
Chaowei Xiao, Dawei Yang, Bo Li, Jia Deng, and Mingyan Liu.
\newblock Meshadv: Adversarial meshes for visual recognition.
\newblock In \emph{Proceedings of the IEEE/CVF Conference on Computer Vision and Pattern Recognition}, pages 6898--6907, 2019.

\bibitem[Xie et~al.(2019)Xie, Zhang, Zhou, Bai, Wang, Ren, and Yuille]{Xie_2019_CVPR}
Cihang Xie, Zhishuai Zhang, Yuyin Zhou, Song Bai, Jianyu Wang, Zhou Ren, and Alan~L. Yuille.
\newblock Improving transferability of adversarial examples with input diversity.
\newblock In \emph{Proceedings of the IEEE/CVF Conference on Computer Vision and Pattern Recognition (CVPR)}, 2019.

\bibitem[Yang et~al.(2022)Yang, Dong, Pang, Su, and Zhu]{yang2022boosting}
Xiao Yang, Yinpeng Dong, Tianyu Pang, Hang Su, and Jun Zhu.
\newblock Boosting transferability of targeted adversarial examples via hierarchical generative networks.
\newblock In \emph{European Conference on Computer Vision}, pages 725--742. Springer, 2022.

\bibitem[Young()]{youngxatlas}
Jonathan Young.
\newblock xatlas, 2021.

\bibitem[Zeng et~al.(2019)Zeng, Liu, Wang, Qiu, Xie, Tai, Tang, and Yuille]{zeng2019adversarial}
Xiaohui Zeng, Chenxi Liu, Yu-Siang Wang, Weichao Qiu, Lingxi Xie, Yu-Wing Tai, Chi-Keung Tang, and Alan~L Yuille.
\newblock Adversarial attacks beyond the image space.
\newblock In \emph{Proceedings of the IEEE/CVF Conference on Computer Vision and Pattern Recognition}, pages 4302--4311, 2019.

\bibitem[Zhang et~al.(2018)Zhang, Isola, Efros, Shechtman, and Wang]{zhang2018unreasonable}
Richard Zhang, Phillip Isola, Alexei~A Efros, Eli Shechtman, and Oliver Wang.
\newblock The unreasonable effectiveness of deep features as a perceptual metric.
\newblock In \emph{Proceedings of the IEEE conference on computer vision and pattern recognition}, pages 586--595, 2018.

\bibitem[Zhao et~al.(2021)Zhao, Liu, and Larson]{zhao2021success}
Zhengyu Zhao, Zhuoran Liu, and Martha Larson.
\newblock On success and simplicity: A second look at transferable targeted attacks.
\newblock \emph{Advances in Neural Information Processing Systems}, 34:\penalty0 6115--6128, 2021.

\bibitem[Zhao et~al.(2019)Zhao, Zheng, Xu, and Wu]{zhao2019object}
Zhong-Qiu Zhao, Peng Zheng, Shou-tao Xu, and Xindong Wu.
\newblock Object detection with deep learning: A review.
\newblock \emph{IEEE transactions on neural networks and learning systems}, 30\penalty0 (11):\penalty0 3212--3232, 2019.

\end{thebibliography}
}
\appendix
\clearpage
\setcounter{page}{1}
\maketitlesupplementary

\section{Physical Implementation Details}
\label{sec:implement}
For the implementation of physical attacks, here we detail the specific process including the used equipment. The whole implementation process could be split into three stages: Preprocess---3D Print---Capture and Test.

\textbf{Preprocess.}
Before 3D printing, there is still a need for some preprocessing of the generated adversarial 3D mesh. This is because we have parameterized the appearance information into the parameters of grid-based NeRF. Hence, it is necessary to convert this information back into the texture maps required for 3D printing. Specifically, to extract the appearance as texture images, we first unwrap the UV coordinates of $\mathcal{M}_{adv}$ using XAtlas~\cite{youngxatlas}. Subsequently, we bake the surface’s color into an image of the texture map corresponding to the UV coordinates, which could be used for the following 3D printing.

\textbf{3D Print.}
Then, we print the 3D adversarial examples generated by TT3D in the form of textured meshes using well-established 3D printers. In our case, we utilize the J850™ Digital Anatomy™ 3D Printer\footnote{\url{https://www.stratasys.com/en/3d-printers/printer-catalog/polyjet/j850-digital-anatomy/}}, a classic 3D printer known for its versatility and precision. This printer can accurately reproduce both the textures and geometries of our adversarial samples, offering a wide range of material choices and color options. Here, we print a total of 20 objects, including 10 against the surrogate model ResNet-101 and 10 against the surrogate DenseNet-121.

\textbf{Capture and Test.} In the last stage, we aim to validate the performance of the 3D adversarial samples generated by TT3D across different viewpoints and backgrounds in the physical world. Thus, as mentioned in the manuscript, we place the 3D adversarial object on the given surface (with different backgrounds \textcolor{blue}{B-1}, \textcolor{blue}{B-2}, and \textcolor{blue}{B-3}) and slowly circle them with a smartphone for about $360^{\circ}$, excluding the bottom part. which lasts approximately 20 seconds per object in each setting, capturing 10 frames per second, resulting in a total of 200 frames. Finally, we could calculate the attack success rate as the final test result, which is determined by the proportion of successful frames.

\section{More Experiments}
\label{sec:more experi}
\subsection{Choice for Epoch}
\label{sec:converg}
\begin{figure}[!t]
    \centering
    \includegraphics[width=\linewidth]{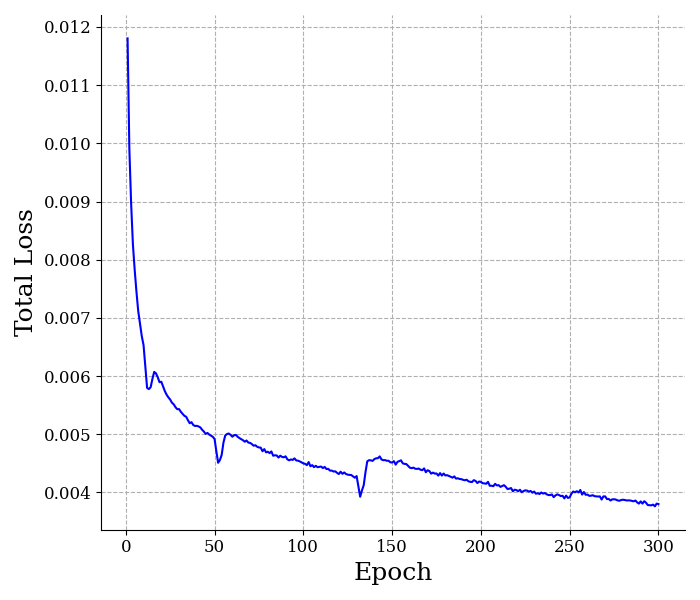}
    \caption{The total loss of TT3D with respect to the epoch against the surrogate model
ResNet-101 with different
training iterations.}
    \label{fig:epoch}
    % \vspace{-3ex}
\end{figure}

To ensure convergence and prevent overfitting, selecting an appropriate epoch for TT3D is crucial. Thus,  following the same settings as described in Section 4.1 of the manuscript, we have visualized the variation curve of the total loss value with respect to the epoch for the attack surrogate model ResNet-101 in \cref{fig:epoch}. The figure reveals that when the epoch reaches the number of 250, the total loss value gradually stabilizes, signifying the achievement of convergence. To prevent overfitting, we have thus chosen 250 as the final epoch for optimization.

\subsection{Evaluation for Naturalness}
\label{sec:naturalness}

\begin{table}[!b]
    % \vspace{-2ex}
    \begin{center}
    \setlength{\tabcolsep}{7pt}
    \resizebox{\linewidth}{!}{
    \begin{tabular}{c|c|c|c|c}
    \hline
      \multicolumn{2}{c|}{\diagbox{Methods}{Metrics}} & SSIM$\uparrow$  & PSNR$\uparrow$  & LPIPS$\downarrow$ \\\hline\hline
    \multicolumn{2}{c|}{Initial}  & 0.9821   &  39.00 & 0.0293\\\hline
     \multicolumn{2}{c|}{Mesh-based} & 0.8259  & 13.79  & 0.1912\\\hline
    \multicolumn{1}{c|}{\multirow{3}{*}{TT3D(RN-101)}} & $\beta=10^{2}$  & 0.8741 & 22.54   & 0.1447 \\\cline{2-5}
    \multicolumn{1}{c|}{} & $\beta=10^{3}$ & 0.9039   & 27.96  & 0.1153 \\\cline{2-5}
  \multicolumn{1}{c|}{}  & $\beta=10^{4}$ & 0.9486  & 33.26   & 0.0874 \\\hline
    \end{tabular}}
    \end{center}
\caption{The results of quantitative naturalness metrics: \textbf{PSNR, SSIM, and LPIPS scores} for the initially reconstructed clean images, adversarial samples generated by the previous mesh-based method, and the corresponding samples produced by our TT3D method under $\beta=10^{2}$, $\beta=10^{3}$, and $\beta=10^{4}$. Higher SSIM and PSNR values indicate better performance, while a lower LPIPS score is preferable.}
    \label{tab:natural}
        \vspace{-3ex}
\end{table}

\begin{figure*}[!t]
    \centering
    \includegraphics[width=\linewidth]{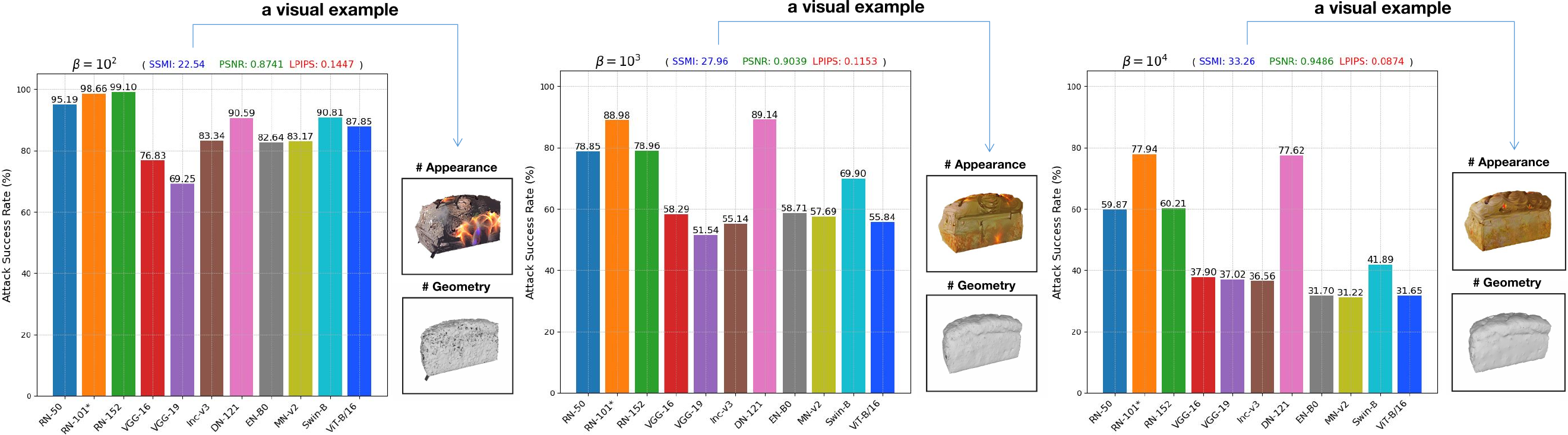}
\caption{The \textbf{attack success rate(\%)} of 3d adversarial examples generated by TT3D under $\beta=10^{2}$, $\beta=10^{3}$, and $\beta=10^{4}$ against ResNet-50 (RN-50), ResNet-101 (RN-101), ResNet-152 (RN-152), VGG-16, VGG-19, Inception-v3 (Inc-v3), DenseNet-121 (DN-121), EfficientNet-B0 (EN-B0), MobileNet-v2 (MN-v2), Swin-B, and VIT-B/16. The 3d adversarial examples are learned against surrogate models ResNet-101. Each bar chart is accompanied by a visual example in the lower right position, showcasing the appearance and geometry of a 3d adversarial sample (attack `bread' to `stove'
) generated by TT3D under the corresponding $\beta$ value. }
    \label{fig:beta_asr}
\end{figure*}
\begin{table*}[!h]
\small
\setlength{\tabcolsep}{3pt}
  \centering
  \resizebox{\linewidth}{!}{
  \begin{tabular}{c|c|c|c|c|c|c|c|c|c|c|c|c}
    \hline
   \multirow{2}{*}{ Source Model } & \multirow{2}{*}{ Background } & \multicolumn{11}{c}{Victim Model} \\ \cline{3-13}
  & &    RN-50 & RN-101 &  RN-152 & VGG-16 & VGG-19 & Inc-v3 &  DN-121 & EN-B0 & MN-v2 & Swin-B & ViT-B/16 \\
    \hline
    \multirow{3}{*}{ResNet-101} 
    & \textcolor{blue}{B-1} & 45.65 & $81.30^{*}$ & 80.20 & \cellcolor{LightCyan}\textbf{31.25} & \cellcolor{LightCyan}\textbf{37.90} & 49.30 & 73.30 & \cellcolor{LightCyan}\textbf{33.45} & 45.30 & \cellcolor{LightCyan}\textbf{71.75}  & 39.25  \\\cline{2-13}
   & \textcolor{blue}{B-2} & 42.35 & $76.45^{*}$ & 62.85 & 21.35 & 23.60 & 47.70 & 61.30 & 23.90 & \cellcolor{LightCyan}\textbf{59.65} & 63.20  &  37.70 \\\cline{2-13}
     &  \textcolor{blue}{B-3} & \cellcolor{LightCyan}\textbf{84.70} & \cellcolor{LightCyan}$\bm{91.55^{*}}$ & \cellcolor{LightCyan}\textbf{89.70} & 29.60 & 32.80 & \cellcolor{LightCyan}\textbf{69.60}  & \cellcolor{LightCyan}\textbf{90.40} &  32.35 & 36.60  & 68.60 & \cellcolor{LightCyan}\textbf{64.15}  \\\cline{2-13}
    \hline
    \multirow{3}{*}{DenseNet-121} 
        & \textcolor{blue}{B-1} & 39.25 & 34.70 & 36.35 & 29.70 & 26.55 &  32.15 & $79.30^{*}$ & 31.20  & 33.75  & 43.20 &  21.20\\\cline{2-13}
           & \textcolor{blue}{B-2} & \cellcolor{LightCyan}\textbf{41.30} & \cellcolor{LightCyan}\textbf{37.65} & \cellcolor{LightCyan}\textbf{39.20} & 35.50 & 41.70 & 34.90 & $83.25^{*}$ & 32.95 & 35.20 & 62.25  & \cellcolor{LightCyan}\textbf{31.85}  \\\cline{2-13}
              & \textcolor{blue}{B-3} & 39.45 & 35.95 & 37.60 & \cellcolor{LightCyan}\textbf{51.55} & \cellcolor{LightCyan}\textbf{42.30} & \cellcolor{LightCyan}\textbf{37.45} & \cellcolor{LightCyan}$\bm{84.50^{*}}$ & \cellcolor{LightCyan}\textbf{34.20} & \cellcolor{LightCyan}\textbf{36.15} & \cellcolor{LightCyan}\textbf{77.30} & 29.80 \\
    \hline
  \end{tabular}}
\caption{The \textbf{attack success rates(\%)} of 3d printed adversarial objects with different backgrounds in the physical world against ResNet-50 (RN-50), ResNet-101 (RN-101), ResNet-152 (RN-152), VGG-16, VGG-19, Inception-v3 (Inc-v3), DenseNet-121 (DN-121), EfficientNet-B0 (EN-B0), MobileNet-v2 (MN-v2), Swin-B, and VIT-B/16. The 3d adversarial objects are learned against the surrogate model ResNet-101 and DenseNet-121 and produced with 3d printing techniques for physical implementation.} 
    \label{tab:physical_transfer}
\end{table*}

To more objectively evaluate the naturalness of 3D adversarial samples generated by TT3D, three metrics for assessing image quality are employed here: Structural Similarity Index~\cite{wang2004image} (SSIM), Peak Signal-to-Noise Ratio (PSNR), and Learned Perceptual Image Patch Similarity~\cite{zhang2018unreasonable} (LPIPS). SSIM evaluates perceptual degradation by analyzing structural, luminance, and contrast changes. PSNR measures pixel-level accuracy between original and distorted images, with higher values indicating better fidelity. LPIPS, leveraging deep learning (vgg used here), evaluates perceptual similarity at the patch level, reflecting the human visual system's response to image variations. Specifically, we compare the above three quantitative metrics of the initially reconstructed clean images, adversarial samples generated by the previous mesh-based method, and the corresponding ones produced by our TT3D method under various $\beta$ values. $\beta$ is the weight of the regularization for naturalness in TT3D. Experimental results 
are listed in \cref{tab:natural}, revealing that our TT3D significantly outperforms the mesh-based method even under different $\beta$ values and exhibits an acceptable decline compared to the clean images, thereby confirming the superior naturalness of TT3D.

\subsection{Effects of $\beta$ }
As mentioned in the manuscript, $\beta$, an adjustable weight for the regularization $\mathcal{R}$ responsible for naturalness, has a significant impact on both naturalness and attack performance. To perform a more comprehensive analysis of $\beta$'s effects, here we measure the attack success rate of TT3D (against the surrogate model ResNet-101) across various black-box classifiers with different $\beta$ values of $10^2$, $10^3$, and $10^4$. Additionally, we provide a visual example of the adversarial samples generated under these settings. Experimental results, as presented in \cref{fig:beta_asr}, where we can see that: 1) \textbf{At $\bm{\beta=10^2}$}, when $\beta$ is relatively low, the success rate across different models is predominantly above 80\%. However, this comes at the cost of a certain degree of naturalness, both in appearance and geometry. 2) \textbf{At $\bm{\beta=10^4}$}, under a strong regularization weight, we still observe considerable success rates, with the lowest being above 30\%. 3) \textbf{At $\bm{\beta=10^3}$}, there is a relative balance between success rate and naturalness, with both metrics demonstrating objectively favorable outcomes. Consequently, this $\beta$ value, i.e., $\beta=10^{3}$ is chosen as the final implementation parameter for TT3D.
\begin{figure*}[!h]
    \centering
    \includegraphics[width=0.94\linewidth]{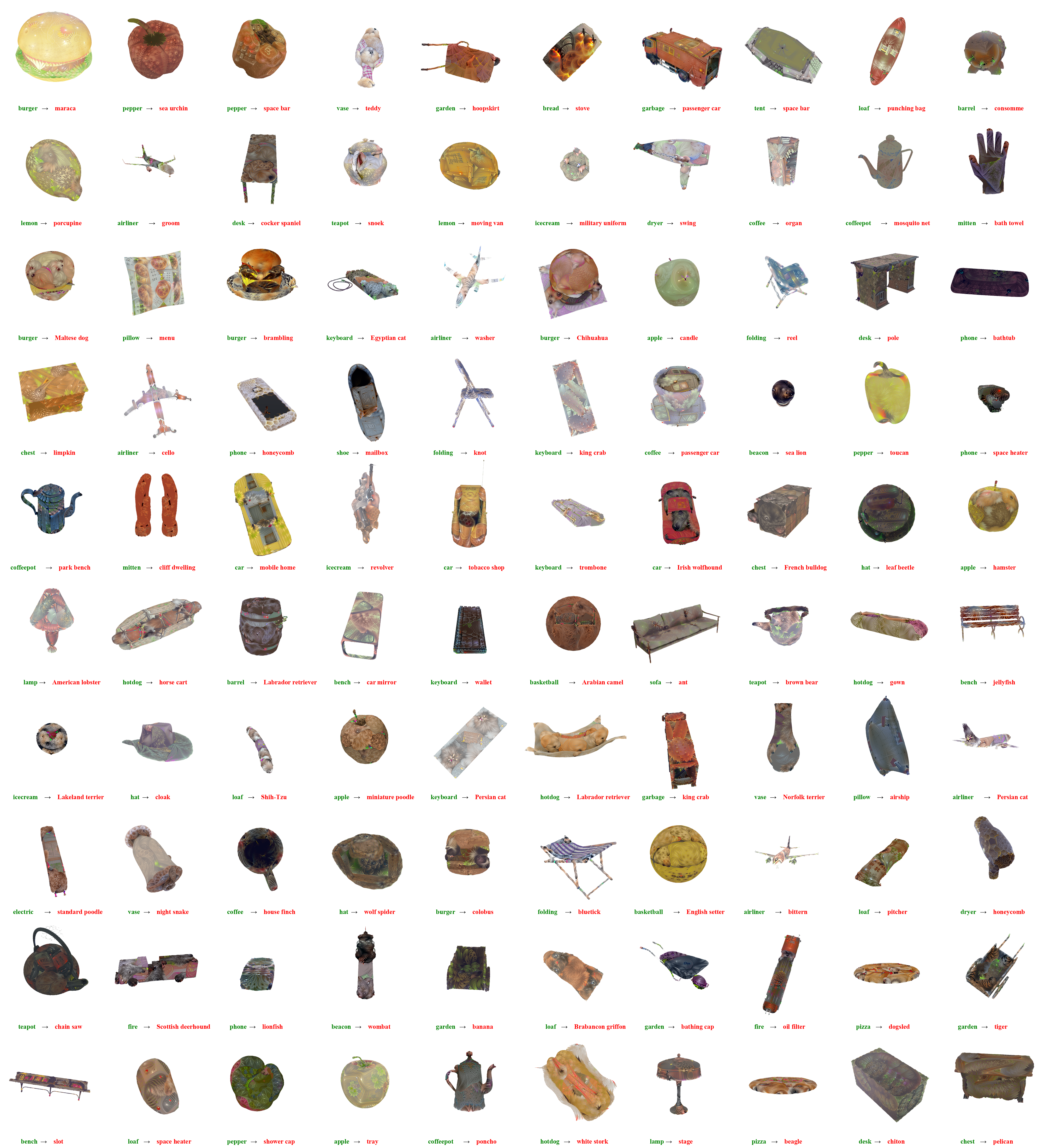}
    \caption{The total of 100 different 3D adversarial objects, generated by TT3D to attack the surrogate models ResNet-101. The above image of each object is captured under a single, random viewpoint, and in the experiments, we capture 100 different viewpoints for calculating the ASR. The green text under each object is the clean label, and the red text is the target label, which is also randomly chosen from the labels in Imagenet. The above images show the diversity of the object classes involved in the attack and the randomness of the target categories.}
    \label{fig:more_examples}
\end{figure*}
\begin{figure*}[!h]
    \centering
    \includegraphics[width=0.94\linewidth]{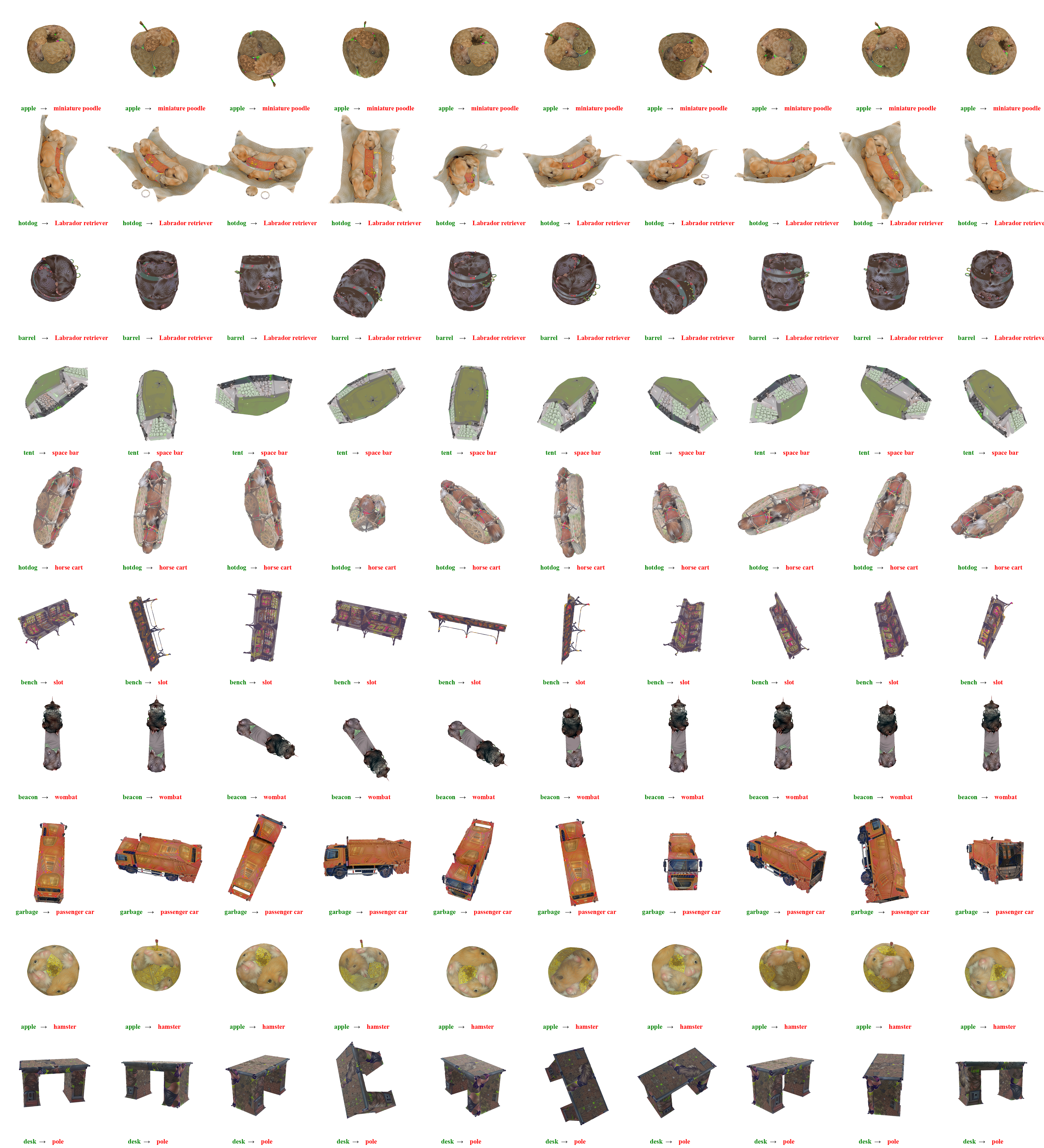}
\caption{The total of 10 randomly selected 3D adversarial objects under 10 random viewpoints, along with their predicted outcomes. This validates the randomness and diversity of our viewpoint selection, as well as the robustness of our TT3D method across varying viewpoints.}
    \label{fig:more_examples_10}
\end{figure*}
\subsection{Transferability in the Physical World}
\label{sec:transfer in phy}
Our TT3D could achieve a transferable targeted attack not only in the digital world but also in the physical world. The specific results in the digital world have been given in the manuscript. Here, to verify the performance in the physical world, we follow the implementation process in \cref{sec:implement} 
and test our 3D-printed adversarial objects' attack success rate against various black-box classifiers in the physical world. As shown in \cref{tab:physical_transfer}, even when confronted with varying backgrounds, our 3D adversarial objects still maintain commendable attack success rates across a spectrum of black-box classifiers, which verifies the robustness of our TT3D.
\section{More examples of TT3D}
In this section, we present more 3D adversarial samples generated by TT3D: (1) \cref{fig:more_examples} shows 100 different 3D adversarial objects, generated by attacking the surrogate model ResNet-101. These images, captured from a single random viewpoint, demonstrate the diversity of the object classes involved in the attack and the randomness of the target categories. (2) \cref{fig:more_examples_10} consists of 10 randomly selected 3D objects from \cref{fig:more_examples}, each depicted from 10 different viewpoints (100 random viewpoints used in the experiments), to illustrate the effectiveness of the 3D adversarial samples under various viewpoints.
\label{more examples}


\end{document}